%% file: arxiv.tex
\documentclass[journal]{IEEEtran}
\usepackage{amsmath,amsfonts}
\usepackage{algorithm}
\usepackage{algpseudocode}
\usepackage{array}
\usepackage{xcolor}
\usepackage{colortbl}  
\usepackage{multirow}
\usepackage{textcomp}
\usepackage{stfloats}
\usepackage{arydshln} 
\usepackage{url}
\usepackage{verbatim}
\usepackage{graphicx}
\usepackage{subcaption}
\usepackage{caption}
\usepackage{booktabs}
\usepackage{amssymb}
\usepackage{bm}

\usepackage[pdfborder={0 0 0}]{hyperref}    
\usepackage{cleveref}    
\usepackage{makecell}
\begin{document}

\title{DSM-Free UAV Visual Localization via Latent-Depth Variable Projection and Structure-Adaptive Coupling}

\author{Yibin Ye,  Xichao Teng,  Shuo Chen, Xiaokai Song, Yuli Sun, Qifeng Yu, Zhang Li,~\IEEEmembership{Member,~IEEE}
\thanks{Yibin Ye, Xichao Teng, Shuo Chen, Xiaokai Song, Qifeng Yu
and Zhang Li are with the College of Aerospace Science and Engineering,
National University of Defense Technology, Changsha 410073, China, and
Hunan Provincial Key Laboratory of Image Measurement and Vision
Navigation, Changsha 410073, China. Yuli Sun is with the College of
Electronic Science and Technology, National University of Defense
Technology, Changsha 410073, China. Corresponding authors: Zhang Li
(E-mail: zhangli\_nudt@163.com).}
}

\markboth{Journal of \LaTeX\ Class Files,~Vol.~14, No.~8, August~2021}%
{Shell \MakeLowercase{\textit{et al.}}: A Sample Article Using IEEEtran.cls for IEEE Journals}


\maketitle
\input{sec/0_abstract}

\input{sec/1_intro}

\input{sec/2_related_work}

\input{sec/3_method_new_s1_sac}
\input{sec/4_experiment_new_s1_sac}
\input{sec/6_conclusion}

 \input{sec/7_appendix_revised_v3}
\bibliographystyle{IEEEtran}
\bibliography{main}

\end{document}

%% file: sec/0_abstract.tex
\begin{abstract}

Visual localization enables unmanned aerial vehicles (UAVs)
to maintain absolute positioning in GNSS-denied environments.
A common approach matches UAV images with geo-tagged
orthographic maps and estimates camera pose using
Perspective-$n$-Point (PnP) with elevations from
Digital Surface Models (DSMs). However, accurate and
up-to-date DSMs are often unavailable in practice.
Orthographic--perspective geometry (OPG) enables DSM-free
localization in scenes with varying elevations, yet
existing methods typically estimate pose indirectly
through intermediate matrix estimation and decomposition.
To address this limitation, we propose
\emph{Latent-Depth Variable Projection} (LDVP), which
treats the unknown scales along perspective viewing
rays as latent depths.
By analytically eliminating these depths and solving
for the horizontal camera center in closed form given
the rotation, LDVP reduces nonlinear estimation to a
rotation-only problem, enabling direct recovery of
metric horizontal position without assuming a planar
scene.
To improve localization robustness across different
scene structures, we further introduce
\emph{Structure-Adaptive Coupling} (SAC), which uses
relative scene structure inferred from monocular depth
estimation to select LDVP when the matched points are not
dominated by a single plane, and Planar-PnP otherwise. Our local observability analysis reveals how scene
elevation variation affects the geometric constraints
of OPG, providing a basis for this adaptive strategy. Extensive experiments show that LDVP consistently
outperforms existing OPG-based DSM-free methods,
while SAC further improves robustness across diverse
scenes and substantially narrows the gap to DSM-based
localization.
Code will be released at
\url{https://github.com/UAV-AVL/LDVP-SAC}.
\end{abstract}


%% file: sec/1_intro.tex
\section{Introduction}
\label{sec}

Unmanned Aerial Vehicles (UAVs) are increasingly used in applications such as inspection, emergency response, and mapping \cite{zhang2025GeoRVLF}. Most outdoor UAV systems rely on the Global Navigation Satellite System (GNSS), together with Inertial Measurement Units (IMUs), for global positioning and navigation. However, GNSS can become unreliable in environments with signal blockage or interference, while inertial navigation inevitably accumulates drift over time \cite{GPS_jamming}. Visual localization therefore plays an important role in maintaining absolute positioning capability and operational autonomy when GNSS is degraded or unavailable \cite{chang2023review}.

Map-based absolute UAV visual localization commonly matches onboard UAV images against geo-tagged orthographic reference maps, such as aerial orthophotos or orthorectified satellite imagery, and recovers the UAV pose from the geometric relationship between the two views \cite{dhaouadi2025ortholoc,Ye_2026_CVPR}. The choice of pose estimation model is closely related to the elevation structure represented by the matched points. When most points lie on or near a common plane, the planar assumption closely matches the scene geometry and provides an effective way to estimate pose without per-point elevation. This condition is often encountered in relatively flat, expansive areas, such as plains, or in scenes dominated by planar structures, including roads and rooftops. Accordingly, planar models have been widely used in UAV localization \cite{chang2023review,wang2024accurate,li2026robust}. In low-altitude flight, however, the validity of this assumption becomes strongly scene dependent. Matched points may span buildings, vegetation, sloped terrain, and other structures at different elevations, producing parallax that cannot be explained by a single plane. A more general solution is to recover the 3D coordinates of the matched map points from a Digital Surface Model (DSM) \cite{dhaouadi2025ortholoc,chen2021real,ye2024coarse} and estimate the camera pose using a Perspective-\(n\)-Point (PnP) solver \cite{Lepetit2009EPnP}.

Although DSM-based PnP can handle general 3D scene structure, obtaining elevation data that meet the requirements of high-accuracy UAV localization remains difficult in practice. High-quality DSM generation usually relies on dedicated stereo photogrammetry, airborne LiDAR, or other specialized mapping procedures. Widely available global elevation products are easier to obtain, but their spatial resolution is often much coarser than that of the 2D reference maps used for visual localization. For example, AW3D30, a widely used global DSM, provides a spatial resolution of approximately 30~m \cite{Nikolakopoulos2020AW3D30}, whereas reference maps used for UAV localization are commonly available at meter-level or finer resolution \cite{dhaouadi2025ortholoc,Ye_2026_CVPR}. This resolution gap makes it difficult to assign sufficiently accurate elevations to individual matched points. Moreover, in rapidly changing urban areas or disaster-affected regions, buildings and local terrain may change after the DSM is acquired, leading to geometric inconsistency between the elevation data and the 2D reference map \cite{deng2026aeromap3d}. Previous studies have also shown that degraded or missing DSM information can noticeably reduce the accuracy of low-altitude UAV localization \cite{Ye_2026_CVPR,Li2026AH}. These limitations motivate methods that can recover accurate pose directly from 2D orthographic maps without requiring DSM elevations.

Orthographic--perspective geometry (OPG)
\cite{Larsson_2021_ICCV} provides a promising way to
address this problem beyond the planar assumption.
By eliminating unknown scene elevations, OPG derives
pose constraints directly from correspondences between
perspective UAV images and orthographic reference maps.
Larsson \emph{et al.} introduced orthographic--perspective
epipolar (OPE) geometry and derived minimal solvers for
the corresponding essential matrix
\cite{Larsson_2021_ICCV}. More recently, Li \emph{et al.}
proposed the Antisymmetric-Homography (AH) constraint
for DSM-free horizontal localization \cite{Li2026AH}.
However, existing OPG-based methods typically recover
pose indirectly through intermediate matrix estimation
and decomposition. Moreover, OPG constraints are
inherently sensitive to scene elevation structure.
Our local observability analysis shows that elevation
variation strengthens pose constraints, whereas
near-planar structures lead to weaker observability.
These challenges highlight the need for both direct
metric pose estimation and adaptation to different
scene structures.

To enable direct metric pose estimation without a DSM,
we propose \emph{Latent-Depth Variable Projection} (LDVP),
a geometric formulation based on OPG. By treating the unknown scale along each perspective
viewing ray as a latent depth, LDVP analytically
eliminates these variables and solves for the horizontal
camera center in closed form given the rotation.
This reduces nonlinear optimization to camera rotation
and enables recovery of the metric horizontal position
without intermediate matrix estimation and decomposition.
Unlike planar methods, LDVP handles elevation variations
without requiring per-point height information.

To improve DSM-free localization across different
scene structures, we further introduce
\emph{Structure-Adaptive Coupling} (SAC).
Since the observability of OPG depends on scene
elevation variation, LDVP may become less reliable
in approximately planar scenes.
For such scenes, we consider planar PnP (Planar-PnP),
which estimates camera pose by assigning a common
elevation to the matched map points.
SAC uses relative scene structure inferred from
monocular depth estimation (MDE) \cite{moge2aerial}
to select LDVP when the matched points are not
dominated by a single plane, and Planar-PnP otherwise. By adapting the pose estimation method to scene
structure, SAC improves localization robustness
across diverse scenes without requiring a DSM.

We evaluate the proposed framework on the AnyVisLoc and OrthoLoC benchmarks \cite{Ye_2026_CVPR,dhaouadi2025ortholoc}, which cover diverse low-altitude environments and orthographic reference maps. Experiments show that LDVP consistently outperforms existing OPG-based DSM-free methods, validating the effectiveness of direct metric pose estimation. The results also confirm the complementary behavior of LDVP and Planar-PnP across different scene structures. By selecting between them accordingly, SAC further improves localization robustness and substantially narrows the performance gap between DSM-free and DSM-based localization. Overall, the main contributions of this work are summarized
as follows:

\begin{itemize}
    \item We propose \emph{Latent-Depth Variable Projection}
    (LDVP), a direct metric formulation for DSM-free UAV
    localization. By analytically eliminating latent ray
    depths and solving for the horizontal camera center
    in closed form given the rotation, LDVP reduces
    nonlinear estimation to camera rotation.

    \item We analyze the local observability of
    orthographic--perspective geometry, revealing how
    scene elevation variation affects pose constraints
    and providing a geometric basis for
    structure-adaptive pose estimation.

    \item We develop \emph{Structure-Adaptive Coupling}
    (SAC), which uses relative scene structure inferred
    from monocular depth estimation to select between
    LDVP and Planar-PnP. Experiments demonstrate improved
    robustness across diverse scenes and a substantially
    reduced gap to DSM-based localization.
\end{itemize}

%% file: sec/2_related_work.tex
\section{Related Work}
\label{sec:related_work}

\subsection{DSM-based UAV Visual Localization}

Map-based absolute UAV visual localization usually starts from establishing correspondences between UAV images and 2D reference maps. Early approaches mainly relied on hand-crafted local features, such as SIFT, SURF, and ORB, which provide efficient feature matching under moderate viewpoint and appearance variations \cite{SIFT,SURF,rublee2011orb}. With the development of deep learning, learned feature representations have gradually become dominant. Sparse feature-based methods, including SuperPoint and LightGlue, improve feature detection and matching robustness through learned representations \cite{detone18superpoint,lightglue}. More recent detector-free and dense matching approaches, such as LoFTR, DKM, and RoMa, further enhance correspondence estimation under large viewpoint and appearance changes \cite{sun2021loftr,dkm,edstedt2024roma}. These advances provide increasingly reliable correspondences for subsequent geometric localization.

 To recover metric UAV poses from image correspondences, many existing approaches introduce external elevation information. DSM-based localization methods lift matched reference pixels to 3D with DSMs and exploit the resulting 2D--3D correspondences for camera pose estimation, typically through Perspective-$n$-Point (PnP) solvers with robust estimation strategies such as Random Sample Consensus (RANSAC) \cite{chen2021real,ye2024coarse,Ye_2026_CVPR,dhaouadi2025ortholoc}. Several studies further improve localization robustness by incorporating gravity-direction constraints \cite{Wu2023uav4l} or inertial attitude priors \cite{li2025imu}. More recently, AeroMap3D combines bare-earth DEMs with OpenStreetMap semantics to filter correspondences affected by unmodeled buildings before DEM-based 2D--3D lifting and PnP-RANSAC pose estimation \cite{deng2026aeromap3d}. Beyond DSM-based approaches, other external 3D scene representations, including Level-of-Detail (LoD) 3D maps \cite{LoD-Loc}, 3D meshes, and Neural Radiance Fields (NeRFs) \cite{zhang2026visual}, have also been explored to provide geometric constraints for visual localization. However, methods in this category rely on accurate and up-to-date external geometric information, which is difficult to acquire and maintain in practical UAV applications. In contrast, this work focuses on metric UAV
localization using 2D reference maps without external elevation or 3D scene models.
\subsection{DSM-free UAV Visual Localization}

To avoid reliance on external elevation information, DSM-free UAV localization methods attempt to recover camera poses from correspondences between UAV images and 2D reference maps. A widely adopted strategy is to assume that the matched scene is approximately planar. Under this assumption, many existing methods estimate
homographies between UAV images and reference maps
\cite{goforth2019gps,he2024leveraging,zhang2025GeoRVLF},
while camera pose can also be estimated by applying
PnP to coplanar 3D points, referred to here as
Planar-PnP \cite{collins2014infinitesimal}. These approaches are effective when the planar assumption closely matches the scene structure, but their accuracy can degrade when matched points span substantial elevation variation \cite{Ye_2026_CVPR}.

Beyond the planar assumption, several studies have explored more general geometric formulations for DSM-free localization. For street-view localization, Sekii \emph{et al.} modeled each 2D reference point in an aerial image as a vertical 3D line with unknown altitude and estimated camera pose from point-to-line constraints \cite{Sekii2013}. Larsson \emph{et al.} developed Orthographic--Perspective Epipolar (OPE) geometry for pose estimation against a 2D reference map without explicit elevation information \cite{Larsson_2021_ICCV}. More recently, Li \emph{et al.} introduced the Antisymmetric-Homography (AH) formulation and its associated essential matrix for metric UAV localization without prior map elevation \cite{Li2026AH}. Unlike planar methods, OPE and AH eliminate the unknown scene elevations and allow pose estimation from orthographic--perspective correspondences with varying elevations. However, these approaches recover pose indirectly through intermediate matrix estimation and decomposition, rather than directly estimating the metric pose. Furthermore, the pose observability of OPG depends on scene elevation variation and may weaken in approximately planar scenes. Existing OPG-based methods do not explicitly account for this scene-dependent limitation.

\subsection{Monocular Depth-guided Visual Localization}

Recent advances in monocular depth estimation (MDE) have substantially improved cross-scene generalization and geometric prediction, making monocular depth an increasingly practical geometric prior for image matching and visual localization \cite{DA3,depthpro,piccinelli2025unidepthv2}. Toft \emph{et al.} used predicted depth to identify planar regions for perspective rectification \cite{toft2020single}, while Huang \emph{et al.} introduced depth-aware homography estimation to handle non-planar scenes \cite{huang2023deep}. Other methods incorporate depth-derived curvature, 3D coordinates, or surface normals into feature representation and matching \cite{wang2023guiding,karpur2024lfm,LiftFeat}. More recently, DECO exploits monocular depth to infer local surface geometry and co-visibility between UAV images and orthographic maps, suppressing geometrically invalid keypoints before feature matching \cite{ye2026deco}. While these studies demonstrate the value of monocular
depth for understanding scene structure, existing
approaches mainly exploit depth cues for feature
matching, geometric filtering, or scene representation. Using relative scene structure to guide pose solver
selection without treating predicted depth as metric
elevation remains less explored in DSM-free UAV
localization.

%% file: sec/3_method_new_s1_sac.tex
\section{Methodology}
\label{sec:method}

\begin{figure*}[t]
    \centering
    \includegraphics[width=\textwidth]{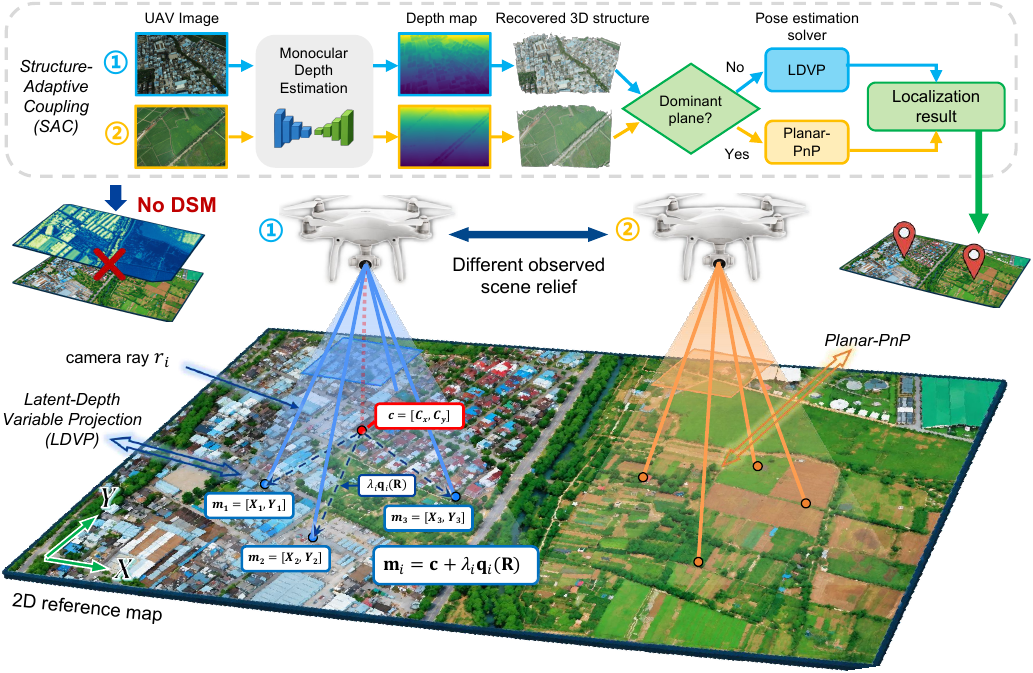}
\caption{
Overview of the proposed DSM-free UAV localization framework.
LDVP directly estimates the metric horizontal camera
position by analytically eliminating the latent depths. SAC uses relative scene structure inferred from monocular depth to adaptively select between LDVP and Planar-PnP, exploiting their complementary strengths across different scene structures.
}
    \label{fig:overview}
\end{figure*}

This section presents the proposed DSM-free UAV localization framework for recovering the metric horizontal camera position from a perspective UAV image and a geo-tagged orthographic reference map. We first formulate \emph{Latent-Depth Variable Projection} (LDVP), which treats the unknown scale along each perspective viewing ray as a latent depth and eliminates it analytically for direct metric pose estimation. LDVP provides a geometric solution for scenes with sufficient relief where planar assumptions become inaccurate. We then analyze the local observability of this formulation and show how scene relief affects the underlying orthographic--perspective geometry. Based on this analysis, we introduce \emph{Structure-Adaptive Coupling} (SAC), which uses relative scene structure inferred from monocular depth to adaptively select between LDVP and Planar-PnP. An overview of the complete framework is shown in Fig.~\ref{fig:overview}.

\subsection{Problem Formulation}
\label{sec:problem_formulation}

Let $\mathbf{I}_u$ denote the perspective UAV image and $\mathbf{I}_r$ a
geo-referenced orthographic reference map. An image matcher provides $N$ point
correspondences
\begin{equation}
    \mathcal{M}
    =
    \left\{
        \left(
            \mathbf{p}_i,
            \mathbf{y}_i
        \right)
    \right\}_{i=1}^{N},
    \label{eq:matches}
\end{equation}
where $\mathbf{p}_i=[u_i,v_i]^{\top}$ is a point in the UAV image and
$\mathbf{y}_i$ is the corresponding point in the reference map. The known
geo-referencing transform $\mathcal{G}(\cdot)$ maps each reference point to
metric horizontal coordinates,
\begin{equation}
    \mathbf{m}_i
    =
    \mathcal{G}(\mathbf{y}_i)
    =
    \begin{bmatrix}
        X_i\\
        Y_i
    \end{bmatrix}
    \in\mathbb{R}^{2}.
    \label{eq:map_xy}
\end{equation}
Without a DSM, $(X_i,Y_i)$ are available while the corresponding elevation
$Z_i$ is unknown.

Given the camera intrinsic matrix
$\mathbf{K}\in\mathbb{R}^{3\times3}$ and the homogeneous UAV-image point
$\widetilde{\mathbf{p}}_i=[u_i,v_i,1]^{\top}$, the unit bearing in the camera
frame is
\begin{equation}
    \mathbf{r}_i
    =
    \frac{
        \mathbf{K}^{-1}\widetilde{\mathbf{p}}_i
    }{
        \left\|
            \mathbf{K}^{-1}\widetilde{\mathbf{p}}_i
        \right\|_2
    }.
    \label{eq:camera_bearing}
\end{equation}

A gravity-aligned world frame is adopted with its $X$--$Y$ plane aligned with
the horizontal map coordinates. Let
$\mathbf{C}=[C_x,C_y,C_z]^{\top}$ be the camera center,
$\mathbf{c}=[C_x,C_y]^{\top}$ its horizontal component, and
$\mathbf{R}\in\mathrm{SO}(3)$ the world-to-camera rotation. The viewing
direction in the world frame is
\begin{equation}
    \mathbf{R}^{\top}\mathbf{r}_i
    =
    \begin{bmatrix}
        \mathbf{q}_i^{\top} & v_{iz}
    \end{bmatrix}^{\top},
    \qquad
    \mathbf{q}_i
    =
    \begin{bmatrix}
        q_{ix} & q_{iy}
    \end{bmatrix}^{\top},
    \label{eq:world_bearing}
\end{equation}
where $\mathbf{q}_i$ and $v_{iz}$ are the horizontal and vertical components
of the world-frame ray.

The corresponding 3D scene point lies on this calibrated ray:
\begin{equation}
    \begin{bmatrix}
        X_i\\
        Y_i\\
        Z_i
    \end{bmatrix}
    =
    \mathbf{C}
    +
    \lambda_i
    \mathbf{R}^{\top}\mathbf{r}_i,
    \qquad
    \lambda_i>0.
    \label{eq:ray_incidence}
\end{equation}
Taking the horizontal components gives
\begin{equation}
    \mathbf{m}_i
    =
    \mathbf{c}
    +
    \lambda_i\mathbf{q}_i(\mathbf{R}).
    \label{eq:ldvp_base}
\end{equation}
As illustrated in Fig.~\ref{fig:overview}, Eq.~\eqref{eq:ldvp_base}
states that the metric map point must lie on the horizontal projection of its
corresponding perspective viewing ray. This projected ray passes through the
horizontal camera center $\mathbf{c}$ with direction $\mathbf{q}_i$, while
$\lambda_i$ represents its unknown scale. Rather than explicitly assigning an
elevation to each correspondence, LDVP captures the unknown scene depth
through this latent ray scale.

\subsection{Latent-Depth Variable Projection}
\label{sec:ldvp}

The main difficulty in Eq.~\eqref{eq:ldvp_base} is that every correspondence
introduces an unknown scale $\lambda_i$. Jointly optimizing these $N$ latent
variables together with the camera pose would unnecessarily enlarge the
nonlinear state. LDVP instead eliminates them analytically through variable
projection.

For fixed $\mathbf{R}$ and $\mathbf{c}$, define the horizontal ray-incidence
residual
\begin{equation}
    \boldsymbol{\epsilon}_i
    =
    \mathbf{m}_i
    -
    \mathbf{c}
    -
    \lambda_i\mathbf{q}_i(\mathbf{R}),
    \label{eq:ldvp_vector_residual}
\end{equation}
which measures the discrepancy between the observed metric map point and the
point predicted on the horizontal camera ray. Let
$\mathbf{d}_i=\mathbf{m}_i-\mathbf{c}$. The optimal latent scale minimizes
$\|\mathbf{d}_i-\lambda_i\mathbf{q}_i\|_2^2$, whose derivative is
\begin{equation}
    \frac{\partial}{\partial\lambda_i}
    \left\|
        \mathbf{d}_i-\lambda_i\mathbf{q}_i
    \right\|_2^2
    =
    -2\mathbf{q}_i^{\top}
    \left(
        \mathbf{d}_i-\lambda_i\mathbf{q}_i
    \right).
    \label{eq:lambda_derivative}
\end{equation}
Setting Eq.~\eqref{eq:lambda_derivative} to zero gives
\begin{equation}
    \lambda_i^{\ast}
    =
    \frac{
        \mathbf{q}_i^{\top}
        (\mathbf{m}_i-\mathbf{c})
    }{
        \mathbf{q}_i^{\top}\mathbf{q}_i
    }.
    \label{eq:lambda_ldvp}
\end{equation}
Thus, each latent depth is eliminated in closed form instead of being included
in the nonlinear pose state. 

Define the unit normal to the horizontal bearing line as
\begin{equation}
    \mathbf{n}_i(\mathbf{R})
    =
    \frac{1}{\|\mathbf{q}_i\|_2}
    \begin{bmatrix}
        -q_{iy}\\
        q_{ix}
    \end{bmatrix}.
    \label{eq:line_normal}
\end{equation}
Substituting Eq.~\eqref{eq:lambda_ldvp} into
Eq.~\eqref{eq:ldvp_vector_residual} gives the signed metric point-to-line
residual
\begin{equation}
    e_i(\mathbf{R},\mathbf{c})
    =
    \mathbf{n}_i^{\top}
    (\mathbf{m}_i-\mathbf{c}),
    \label{eq:ldvp_residual}
\end{equation}
which satisfies
\begin{equation}
    \min_{\lambda_i\in\mathbb{R}}
    \left\|
        \mathbf{m}_i-\mathbf{c}
        -\lambda_i\mathbf{q}_i
    \right\|_2^2
    =
    e_i^2.
    \label{eq:exact_metric_residual}
\end{equation}
Eq.~\eqref{eq:exact_metric_residual} shows that the LDVP residual is the
metric point-to-line error obtained after eliminating the unknown ray scale.
It therefore follows directly from the ray-incidence geometry rather than
from an algebraic approximation.

To further reduce the variables in the outer optimization, the horizontal
camera center is profiled for each rotation:
\begin{equation}
    \mathbf{c}^{\ast}(\mathbf{R})
    =
    \mathop{\arg\min}_{\mathbf{c}}
    \sum_{i=1}^{N}
    \rho
    \left(
        e_i(\mathbf{R},\mathbf{c})
    \right),
    \label{eq:center_profile}
\end{equation}
where $\rho(\cdot)$ is a robust loss. Because
Eq.~\eqref{eq:ldvp_residual} is affine in $\mathbf{c}$ for fixed
$\mathbf{R}$, this inner problem is only a two-dimensional robust regression.
The outer nonlinear problem therefore contains only camera rotation:
\begin{equation}
    \mathbf{R}^{\ast}
    =
    \mathop{\arg\min}_{\mathbf{R}\in\mathrm{SO}(3)}
    \sum_{i=1}^{N}
    \rho
    \left(
        e_i
        \left(
            \mathbf{R},
            \mathbf{c}^{\ast}(\mathbf{R})
        \right)
    \right).
    \label{eq:ldvp_profiled_objective}
\end{equation}
The metric horizontal camera center is finally recovered as
$\mathbf{c}^{\ast}(\mathbf{R}^{\ast})$. Thus, LDVP eliminates the
per-correspondence latent depths and profiles the horizontal camera center,
leaving camera rotation as the only variable in the nonlinear outer
optimization. Detailed derivations of the variable-projection formulation are
provided in the Appendix.

To obtain a robust estimate in the presence of incorrect matches, we
embed the profiled formulation in an M-estimator SAmple Consensus (MSAC) framework \cite{torr2000mlesac}.
After eliminating the latent scale, each correspondence provides one scalar
incidence constraint on the five pose degrees of freedom
$(\mathbf{R},\mathbf{c})$. We therefore use five correspondences to generate
each pose hypothesis. Hypotheses with insufficient support from correspondences satisfying the
positive-depth condition $\lambda_i^{\ast}>0$ are discarded before
consensus evaluation. The remaining
hypotheses are scored over all correspondences using the truncated metric
LDVP residual
\begin{equation}
    E_{\mathrm{MSAC}}(\mathbf{R},\mathbf{c})
    =
    \sum_{i=1}^{N}
    \min
    \left(
        e_i^2(\mathbf{R},\mathbf{c}),
        \tau_{\mathrm{L}}^2
    \right),
    \label{eq:ldvp_msac}
\end{equation}
where $\tau_{\mathrm{L}}$ is the metric inlier threshold. The hypothesis with
the lowest MSAC cost defines the consensus set, which is finally refined using
the profiled LDVP objective.

\subsection{Relation of LDVP to OPE and AH}
\label{sec:relation_ope_ah}

LDVP shares the same underlying orthographic--perspective geometry as OPE
\cite{Larsson_2021_ICCV} and AH \cite{Li2026AH}. We establish their relation
below to support the observability analysis in Sec.~\ref{sec:scene_coupling}.

Write the rotation as
\begin{equation}
    \mathbf{R}
    =
    \begin{bmatrix}
        \mathbf{a}_1 &
        \mathbf{a}_2 &
        \mathbf{a}_3
    \end{bmatrix}.
\end{equation}
In OPE \cite{Larsson_2021_ICCV}, a perspective bearing and an orthographic
map point satisfy
\begin{equation}
    \overline{\mathbf{m}}_i^{\top}
    \mathbf{E}_{\mathrm{OP}}
    \mathbf{r}_i
    =
    0,
    \qquad
    \overline{\mathbf{m}}_i
    =
    \begin{bmatrix}
        X_i & Y_i & 1
    \end{bmatrix}^{\top},
    \label{eq:ope_constraint}
\end{equation}
where, under our pose convention, the OPE matrix can be written as
\begin{equation}
    \mathbf{E}_{\mathrm{OP}}
    =
    \begin{bmatrix}
        -\mathbf{a}_2^{\top} \\
        \mathbf{a}_1^{\top} \\
        C_x\mathbf{a}_2^{\top}
        -
        C_y\mathbf{a}_1^{\top}
    \end{bmatrix}.
    \label{eq:ope_matrix_pose}
\end{equation}

AH \cite{Li2026AH} expresses the same geometry as
\begin{equation}
    \mathbf{r}_i^{\top}
    \mathbf{E}_{\mathrm{AH}}
    \overline{\mathbf{m}}_i
    =
    0,
    \qquad
    \mathbf{E}_{\mathrm{AH}}
    =
    [\mathbf{a}_3]_{\times}
    \begin{bmatrix}
        \mathbf{a}_1 &
        \mathbf{a}_2 &
        \mathbf{t}
    \end{bmatrix},
    \label{eq:ah_constraint}
\end{equation}
with $\mathbf{t}=-\mathbf{R}\mathbf{C}$.

Using the orthonormality of the rotation columns gives
\begin{equation}
    \mathbf{E}_{\mathrm{AH}}
    =
    -\mathbf{E}_{\mathrm{OP}}^{\top},
\end{equation}
and the corresponding residuals satisfy
\begin{equation}
    \mathbf{r}_i^{\top}
    \mathbf{E}_{\mathrm{AH}}
    \overline{\mathbf{m}}_i
    =
    -
    \overline{\mathbf{m}}_i^{\top}
    \mathbf{E}_{\mathrm{OP}}
    \mathbf{r}_i
    =
    -
    \|\mathbf{q}_i\|_2
    e_i(\mathbf{R},\mathbf{c}).
    \label{eq:ope_ah_ldvp_relation_main}
\end{equation}

Therefore, LDVP, OPE, and AH encode the same noiseless incidence constraint,
up to the sign and transpose conventions of their matrix parameterizations.
This equivalence shows that the observability properties analyzed below arise
from the underlying orthographic--perspective geometry rather than from the
specific LDVP parameterization. The complete derivation of
Eq.~\eqref{eq:ope_ah_ldvp_relation_main} is provided in the Appendix.

\subsection{Scene-Dependent Observability Analysis}
\label{sec:scene_coupling}

The pose observability of LDVP depends on the elevation structure of the
matched scene. To characterize this dependence, we analyze the local Jacobian
of the LDVP incidence constraint and examine how scene relief affects the
geometric constraints on camera rotation and horizontal translation.

Consider the unnormalized incidence residual
\begin{equation}
    g_i
    =
    \|\mathbf{q}_i\|_2 e_i
    =
    -q_{iy}(X_i-C_x)
    +
    q_{ix}(Y_i-C_y),
    \label{eq:unnormalized_residual}
\end{equation}
and define the relative scene height as
\begin{equation}
    h_i
    =
    Z_i-C_z
    =
    \lambda_i v_{iz}.
    \label{eq:relative_height}
\end{equation}
For a small pose perturbation
$\delta\boldsymbol{\omega}
=[\omega_x,\omega_y,\omega_z]^{\top}$
and
$\delta\mathbf{c}
=[\delta C_x,\delta C_y]^{\top}$,
the first-order variation of Eq.~\eqref{eq:unnormalized_residual} is
\begin{equation}
\begin{split}
    \delta g_i
    ={}&
    q_{iy}\delta C_x
    -
    q_{ix}\delta C_y
    \\
    &+
    h_i
    \left(
        q_{ix}\omega_x
        +
        q_{iy}\omega_y
    \right)
    \\
    &-
    \lambda_i
    \left(
        q_{ix}^{2}+q_{iy}^{2}
    \right)
    \omega_z.
\end{split}
\label{eq:linearized_residual}
\end{equation}

If all matched points lie on a common horizontal plane, then
$h_i=H$ for all $i$, and the corresponding Jacobian columns satisfy
\begin{equation}
    \mathbf{J}_{\omega_x}
    =
    -H\mathbf{J}_{C_y},
    \qquad
    \mathbf{J}_{\omega_y}
    =
    H\mathbf{J}_{C_x}.
    \label{eq:planar_jacobian_dependency}
\end{equation}
Thus, the two tilt directions become first-order coupled with horizontal
translation and no longer provide independent rotational constraints after
the horizontal camera center is profiled out. This rotation--translation
coupling reduces the local pose observability in the planar limit.

Elevation variation breaks this dependency. Writing
$h_i=H+\Delta h_i$ gives
\begin{equation}
    \mathbf{J}_{\omega_x}
    =
    -H\mathbf{J}_{C_y}
    +
    \Delta\mathbf{h}\odot\mathbf{q}_x,
    \qquad
    \mathbf{J}_{\omega_y}
    =
    H\mathbf{J}_{C_x}
    +
    \Delta\mathbf{h}\odot\mathbf{q}_y,
    \label{eq:relief_breaks_dependency}
\end{equation}
where $\odot$ denotes elementwise multiplication. The terms induced by
$\Delta h_i$ cannot, in general, be absorbed by a single horizontal
translation and therefore introduce additional rotational constraints.
Elevation variation thus breaks the planar
rotation--translation coupling and improves
the local observability of LDVP.

The above analysis is not specific to the LDVP parameterization. As shown in
Eq.~\eqref{eq:ope_ah_ldvp_relation_main}, the OPE and AH residuals differ from
the LDVP residual only by a nonzero scale and sign at a noiseless solution,
and their local Jacobians therefore share the same nullspace. The planar
observability limitation is thus a property of the underlying
orthographic--perspective geometry shared by LDVP, OPE, and AH, rather than a
limitation specific to LDVP. The derivation of
Eq.~\eqref{eq:linearized_residual} and the Jacobian-nullspace relation are
provided in the Appendix.

\subsection{Structure-Adaptive Coupling}
\label{sec:sac}

The analysis in Sec.~\ref{sec:scene_coupling}
shows that the pose observability of OPG weakens
in the planar limit. This motivates SAC to adaptively select between LDVP and Planar-PnP based on
the structure of the matched points.

SAC uses monocular depth only to infer the relative structure of the matched
points and determine whether they are dominated by a planar structure. Let
\begin{equation}
    \widehat{\mathbf{D}}
    =
    \mathcal{F}_{\mathrm{MDE}}(\mathbf{I}_u)
\end{equation}
denote the monocular depth prediction. Each matched UAV-image point
$\mathbf{p}_i$ with a valid depth prediction is back-projected to the camera
frame as
\begin{equation}
    \widehat{\mathbf{X}}_i
    =
    \widehat d(\mathbf{p}_i)
    \mathbf{K}^{-1}
    \widetilde{\mathbf{p}}_i .
    \label{eq:mde_backprojection}
\end{equation}
These 3D points are used only for relative structural analysis; the predicted
depths are not used as metric scene elevations in either pose solver.

We fit a dominant plane to the valid back-projected matched points using
three-point RANSAC. To remove the arbitrary global scale of monocular depth, we normalize the
back-projected points by the median valid predicted depth:
\begin{equation}
    \widetilde{\mathbf{X}}_i
    =
    \frac{\widehat{\mathbf{X}}_i}{\widehat d_{\mathrm{med}}}.
\end{equation}
The normalized point-to-plane residual is
\begin{equation}
    \bar r_i^{\pi}
    =
    \left|
        \widehat{\mathbf{n}}_{\pi}^{\top}
        \widetilde{\mathbf{X}}_i+b_{\pi}
    \right|,
\end{equation}
and a point is considered an inlier if
$\bar r_i^{\pi}<\tau_{\pi}$, where $\tau_{\pi}$ is a fixed dimensionless
threshold. Let $N_1$ denote the number of dominant-plane inliers and
$N_{\mathrm{valid}}$ the number of valid back-projected matched points.
We define the dominant-plane support ratio as
\begin{equation}
    S_1
    =
    \frac{N_1}{N_{\mathrm{valid}}},
    \label{eq:sac_s1}
\end{equation}
which provides a practical measure of scene planarity
without requiring the matched points to share a common
elevation.

LDVP remains the default solver because it does not
require a planar scene assumption.
The solver is selected according to the
dominant-plane support ratio:
\begin{equation}
    \mathcal{B}
    =
    \begin{cases}
        \mathrm{L}, & S_1 < \tau_S, \\
        \mathrm{P}, & S_1 \ge \tau_S,
    \end{cases}
    \label{eq:sac_rule}
\end{equation}
where $\mathrm{L}$ and $\mathrm{P}$ denote LDVP and
Planar-PnP, respectively, and $\tau_S$ is the
planar-support threshold, set to $0.5$ by default.
LDVP is selected when fewer than half of the valid
matched points support the dominant plane, while
Planar-PnP is selected otherwise.
\subsection{Planar-PnP}
\label{sec:planar_pnp}

When SAC detects a dominant planar structure, the corresponding map points are
lifted to a virtual zero-elevation plane:
\begin{equation}
    \mathbf{P}_i^{\pi}
    =
    \begin{bmatrix}
        X_i & Y_i & 0
    \end{bmatrix}^{\top}.
    \label{eq:planar_pnp_point}
\end{equation}
The resulting 2D--3D correspondences are used for pose estimation with
PnP-RANSAC.

The choice of zero elevation does not require the actual plane elevation to be
known. If the physical plane has elevation $Z_{\pi}$, define
\begin{equation}
    \mathbf{C}^{\pi}
    =
    \mathbf{C}
    -
    Z_{\pi}\mathbf{e}_z .
\end{equation}
Then
\begin{equation}
    \begin{bmatrix}
        X_i\\
        Y_i\\
        Z_{\pi}
    \end{bmatrix}
    -
    \mathbf{C}
    =
    \mathbf{P}_i^{\pi}
    -
    \mathbf{C}^{\pi}.
    \label{eq:planar_gauge}
\end{equation}
Thus, setting the plane elevation to zero only shifts the vertical reference
and preserves the camera rotation and horizontal position. The complete framework is summarized in
Algorithm~\ref{alg:ldvp_sac}.

\begin{algorithm}[h]
\caption{Structure-Adaptive DSM-Free UAV Localization}
\label{alg:ldvp_sac}
\begin{algorithmic}[1]
\Require UAV image $\mathbf{I}_u$, orthographic map $\mathbf{I}_r$,
camera intrinsics $\mathbf{K}$, correspondences $\mathcal{M}$,
geo-referencing transform $\mathcal{G}$, planar-support threshold $\tau_S$
\Ensure Estimated camera pose

\State Predict monocular depth
$\widehat{\mathbf{D}} \gets \mathcal{F}_{\mathrm{MDE}}(\mathbf{I}_u)$

\State Back-project the matched UAV-image points with valid depth predictions
using $\widehat{\mathbf{D}}$ and $\mathbf{K}$

\State Normalize the back-projected points by the
median valid predicted depth

\State Fit the dominant plane to the normalized matched points
using RANSAC

\State Compute the dominant-plane support ratio
$S_1 \gets N_1 / N_{\mathrm{valid}}$

\If{$S_1 < \tau_S$}
    \State Estimate the pose using LDVP
\Else
    \State Lift the corresponding map points to the virtual plane
    $\mathbf{P}_i^{\pi} = [X_i, Y_i, 0]^{\top}$
    \State Estimate the pose using Planar-PnP
\EndIf

\State \Return estimated camera pose
\end{algorithmic}
\end{algorithm}

%% file: sec/4_experiment_new_s1_sac.tex
\section{Experiments}
\label{sec:experiments}

\subsection{Experimental Setup}
\label{sec:experimental_setup}

\noindent\textbf{Evaluation protocol.}
We evaluate the proposed method through synthetic and real-world experiments.
The synthetic experiments isolate the effect of scene relief to examine how
elevation variation affects LDVP and the complementary behavior of LDVP and
Planar-PnP. The real-world experiments evaluate localization accuracy and
robustness across diverse scene structures and viewing conditions on
AnyVisLoc \cite{Ye_2026_CVPR} and OrthoLoC
\cite{dhaouadi2025ortholoc}.

For all methods, localization accuracy is evaluated using the horizontal
camera-position error
\begin{equation}
    e_{\mathrm{pos}}
    =
    \left\|
        \widehat{\mathbf{c}}-\mathbf{c}_{\mathrm{gt}}
    \right\|_2,
    \label{eq:position_error}
\end{equation}
where $\widehat{\mathbf{c}}$ and $\mathbf{c}_{\mathrm{gt}}$ denote the
estimated and ground-truth horizontal camera centers, respectively.
We report the
median position error and the localization accuracy $A@\tau$, defined
as the percentage of UAV images whose horizontal position error
does not exceed $\tau$ meters ($\tau\in\{1,3,5\}$).

\noindent\textbf{Compared methods.}
We compare the proposed LDVP and LDVP+SAC with four representative
localization methods. DSM-PnP uses reference-map elevations to construct metric 2D--3D correspondences and estimates the camera pose using PnP-RANSAC \cite{Ye_2026_CVPR}. With meter- to decimeter-level DSM spatial resolution, it serves as a reliable reference for evaluating the impact of unavailable elevation information. Planar-PnP assigns a common zero elevation to the matched map points and
provides a direct DSM-free solution under the planar assumption. OPE \cite{Larsson_2021_ICCV} and AH \cite{Li2026AH} are representative
OPG-based DSM-free methods. We use their respective robust estimation procedures in all experiments. All methods are evaluated using the same UAV-image--reference-map correspondences within
each test instance. DSM elevations are provided only to DSM-PnP; the
remaining methods operate without DSM information.
\begin{figure*}[!b]
    \centering
    \subfloat[Example smooth terrain with $\Delta h=16$~m.
    \label{fig:synthetic_terrain}]{
        \includegraphics[width=0.305\textwidth]
        {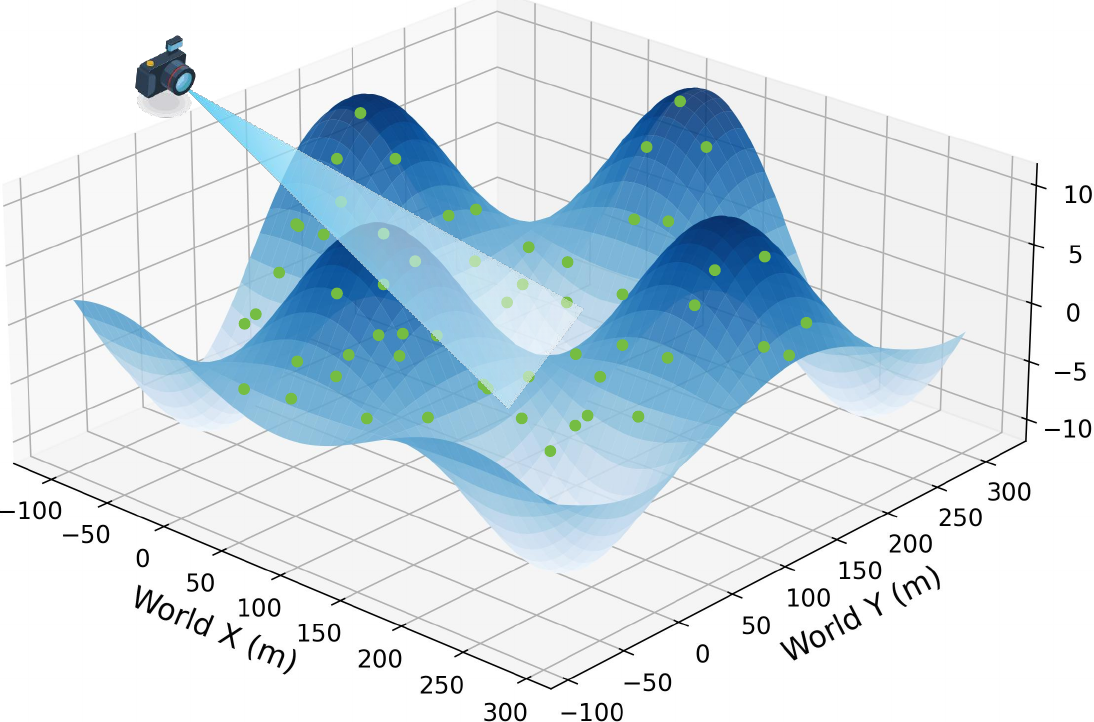}
    }
    \hfill
    \subfloat[Localization error vs. scene relief.
    \label{fig:synthetic_error}]{
        \includegraphics[width=0.305\textwidth]
        {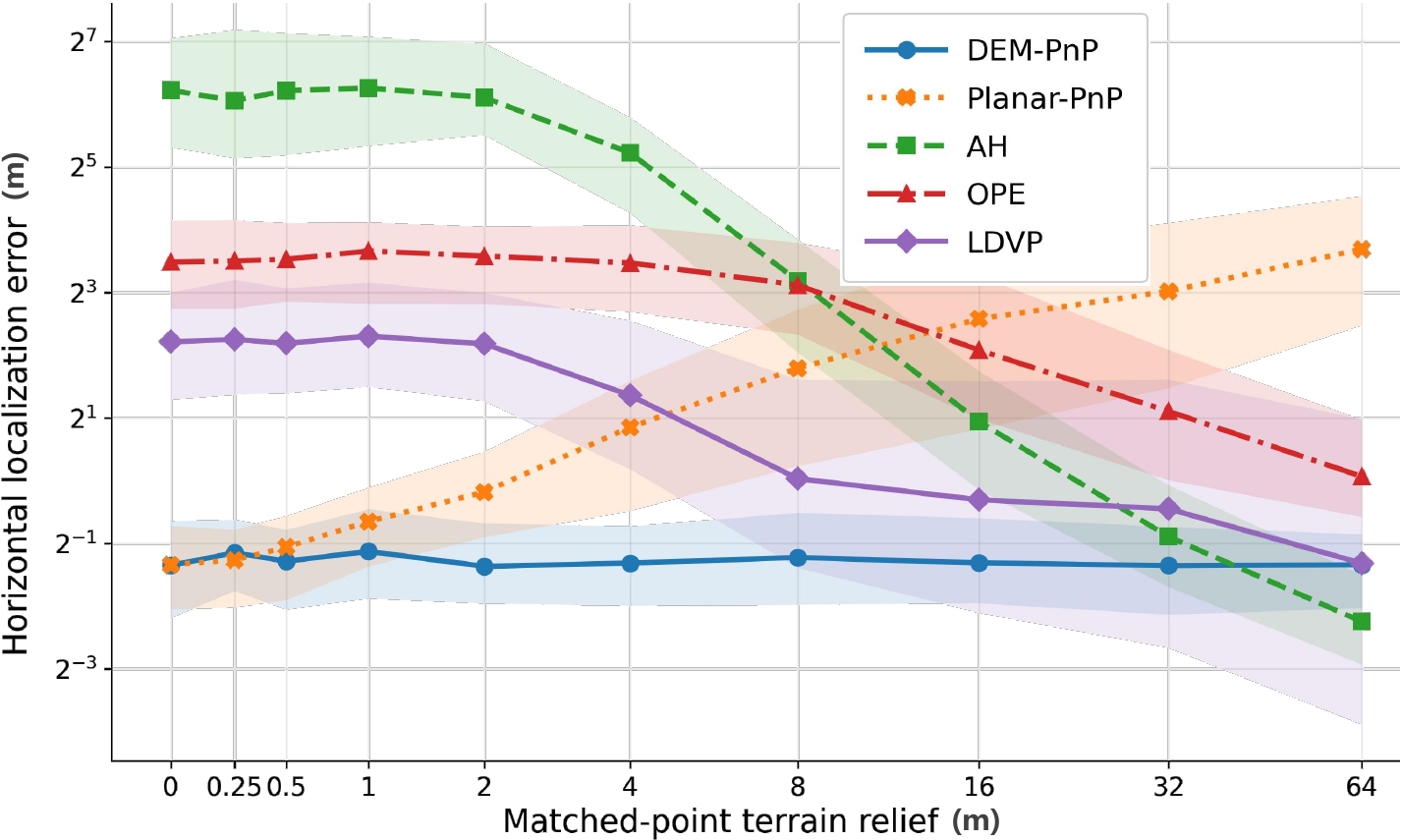}
    }
    \hfill
    \subfloat[$A@1$ vs. scene relief.
    \label{fig:synthetic_recall}]{
        \includegraphics[width=0.305\textwidth]
        {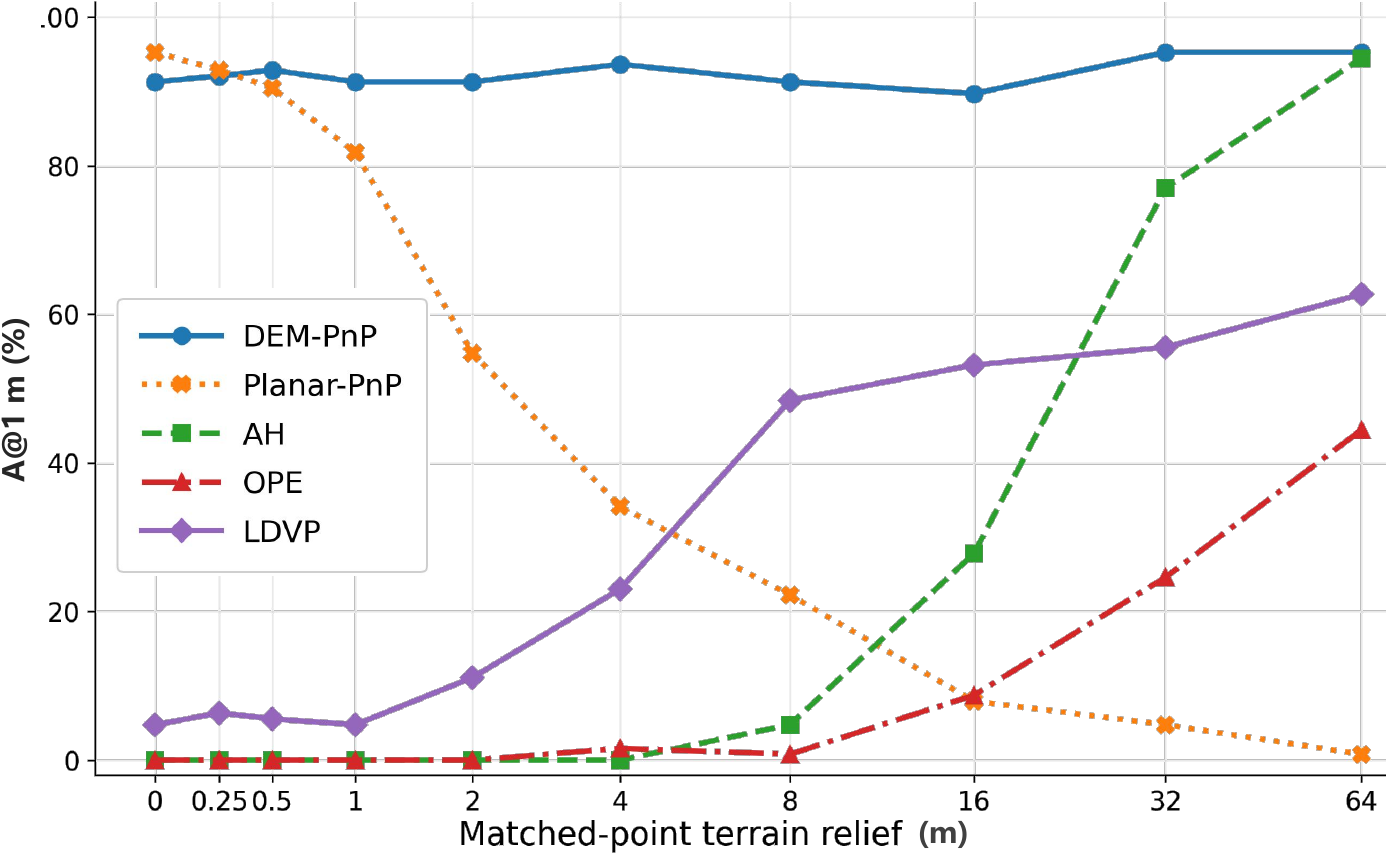}
    }
    \caption{
    Synthetic evaluation under the smooth sinusoidal terrain model.
    The x-axis in (b) and (c) denotes matched-point
relief ($P_{90}-P_{10}$, in meters), where
$P_{90}$ and $P_{10}$ are the 90th and 10th
percentiles of matched-point elevations.
}
    \label{fig:synthetic_relief}
\end{figure*}

\noindent\textbf{Implementation details.}
The experiments are conducted on Ubuntu 22.04 using Python 3.10 and
PyTorch, with an NVIDIA GeForce RTX 4090 GPU and an Intel Xeon
Platinum 8370C CPU. Unless otherwise specified, UAV-image--reference-map correspondences are obtained using RoMa \cite{roma}, a strong dense matching method that provides reliable correspondences under large viewpoint changes. Monocular depth maps used by SAC are predicted using MoGe2-Aerial \cite{moge2aerial}, which is specifically adapted to aerial imagery and is
therefore well suited for inferring the relative scene structure used by SAC. DSM-PnP uses P3P-based RANSAC, while Planar-PnP uses IPPE-based RANSAC,
both implemented with OpenCV. For LDVP, five-correspondence hypotheses are evaluated within MSAC using a
\(5\)~m point-to-line residual threshold. The initial rotation search uses
angular steps of \(30^\circ\), \(45^\circ\), and \(45^\circ\) for tilt, azimuth,
and roll, respectively, followed by local rotation refinement. For SAC, the dominant plane is
estimated using three-point RANSAC with 2000 iterations and a dimensionless inlier threshold of $\tau_{\pi}=0.003$. The planar-support threshold is set to \(\tau_S=0.5\) in the main
experiments.

\subsection{Synthetic Experiments}
\label{sec:synthetic_experiments}

We conduct controlled synthetic experiments to isolate the effect of scene
relief on pose estimation. In particular, we examine whether increasing
elevation variation improves LDVP while making the planar assumption less
accurate, as predicted by the analysis in Sec.~\ref{sec:scene_coupling}.

As shown in Fig.~\ref{fig:synthetic_relief}(a), a continuous elevation field is constructed to simulate scenes with
different levels of height variation. The terrain is defined as
\begin{equation}
g(x,y)
=
\sin\left(\frac{2\pi x}{L}\right)
\cos\left(\frac{2\pi y}{L}\right),
\label{eq:synthetic_base_terrain}
\end{equation}
where \(L=300\)~m controls the spatial wavelength to produce moderate terrain variations. Its elevation is normalized as

\begin{equation}
Z(x,y;\Delta h)
=
\Delta h
\frac{
g(x,y)-P_{50}(g)
}{
P_{90}(g)-P_{10}(g)
},
\label{eq:synthetic_terrain}
\end{equation}
where $P_q(\cdot)$ denotes the $q$-th percentile and $\Delta h$ specifies the
scene relief, measured by $P_{90}-P_{10}$. We evaluate
$\Delta h\in\{0,0.25,0.5,1,2,4,8,16,32,64\}$~m.

Synthetic UAV images have a resolution of $1024\times768$ pixels and a focal
length of 800~pixels. We evaluate six flight altitudes
$\{50,100,150,200,250,300\}$~m and seven camera down-tilt angles
$\{30^\circ,40^\circ,\ldots,90^\circ\}$. Each configuration is repeated three
times with 200 correspondences. Independent Gaussian noise with a standard
deviation of 1~pixel is added to both UAV-image and reference-map points. The
same camera configuration, correspondence layout, and noise realization are
reused across relief levels for paired comparison. Results are pooled over all
flight altitudes and camera orientations.

\begin{table*}[!b]
    \centering
    \caption{Localization results on AnyVisLoc and OrthoLoC using RoMa
    correspondences. Best and second-best DSM-free results are shown in
    \textbf{bold} and \underline{underline}, respectively. DSM-PnP is excluded
    from the ranking.}
    \label{tab:real_world_main}
    \setlength{\tabcolsep}{6pt}
    \renewcommand{\arraystretch}{1.12}
    \begin{tabular}{lrrrrrrrr}
        \toprule
        \multirow{2}{*}{Method}
        & \multicolumn{4}{c}{AnyVisLoc}
        & \multicolumn{4}{c}{OrthoLoC} \\
        \cmidrule(lr){2-5}\cmidrule(lr){6-9}
        & $A@1$ (\%) & $A@3$ (\%) & $A@5$ (\%) & Med. (m)
        & $A@1$ (\%) & $A@3$ (\%) & $A@5$ (\%) & Med. (m) \\
        \midrule
        \rowcolor{gray!15}
        DSM-PnP & 82.22 & 96.74 & 97.20 & 0.46
                         & 72.48 & 84.09 & 86.55 & 0.46 \\
        Planar-PnP & 16.50 & 50.72 & 66.17 & 2.93
                   & 17.49 & 53.66 & \underline{69.64} & 2.72 \\
        AH & 28.01 & 54.87 & 64.99 & 2.44
           & 19.16 & 35.37 & 41.68 & 11.36 \\
        OPE & 41.37 & 70.76 & 80.91 & 1.37
            & 36.73 & \underline{59.66} & 69.28 & 1.83 \\
        LDVP (ours) & \underline{48.59} & \underline{75.41}
                    & \underline{82.87} & \underline{1.05}
                    & \textbf{39.89} & 59.28 & 66.23 & \underline{1.69} \\
        LDVP+SAC (ours) & \textbf{51.35} & \textbf{81.51}
                        & \textbf{88.36} & \textbf{0.97}
                        & \underline{39.34} & \textbf{62.05}
                        & \textbf{70.33} & \textbf{1.66} \\
        \bottomrule
    \end{tabular}
\end{table*}

\begin{figure*}[b]
    \centering
    \includegraphics[width=\textwidth]{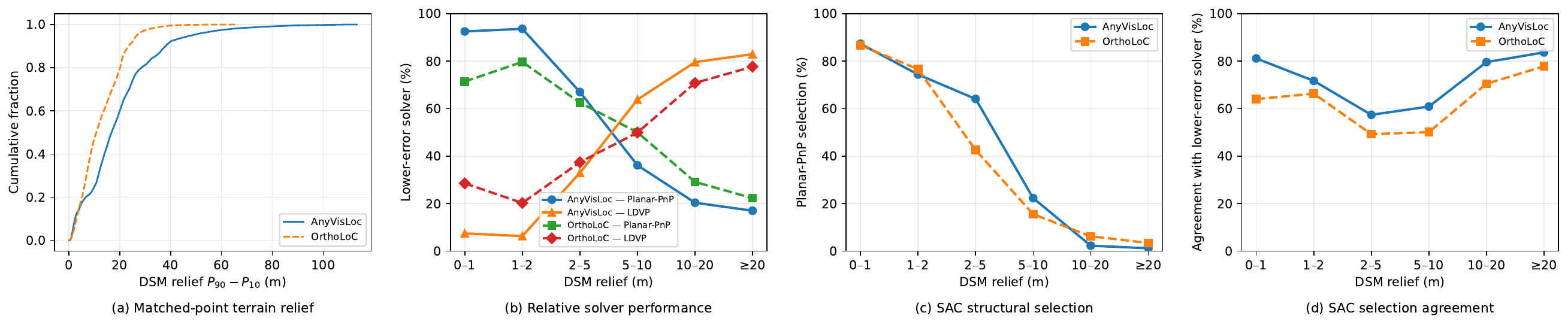}
    \caption{
    Structural analysis of Planar-PnP, LDVP, and SAC on AnyVisLoc and
    OrthoLoC using RoMa correspondences.
    (a) Cumulative distribution of the true matched-point DSM relief,
    measured by $P_{90}-P_{10}$.
    (b) Fraction of queries for which Planar-PnP or LDVP yields the lower
    localization error in each relief interval.
    (c) Fraction of queries for which SAC selects Planar-PnP.
    (d) Agreement between the SAC-selected branch and the lower-error solver.
    }
    \label{fig:real_dsm_structure}
\end{figure*}

As shown in Fig.~\ref{fig:synthetic_relief}(b) and Fig.~\ref{fig:synthetic_relief}(c), DSM-PnP remains accurate across
all relief levels because the true scene elevations are available.
Planar-PnP performs well in near-planar cases, where the planar assumption
closely matches the scene structure, but its accuracy decreases as scene
relief increases. In contrast, the OPG-based DSM-free methods, including LDVP, OPE, and AH,
benefit from increasing scene relief, consistent with the additional geometric constraints identified in Sec.~\ref{sec:scene_coupling}. Among these methods, LDVP consistently outperforms OPE and AH across the tested relief levels, supporting the effectiveness of its direct metric pose formulation under the same underlying OPG. The opposite trends of LDVP and Planar-PnP further confirm their complementary behavior and motivate the structure-adaptive selection used by SAC.
\subsection{Real-World Experiments}
\label{sec:real_world_experiments}

\subsubsection{Datasets}
\label{sec:real_world_datasets}

\noindent\textbf{AnyVisLoc} \cite{Ye_2026_CVPR} comprises 20,077 UAV images collected across
24 diverse urban and natural environments. The dataset covers flight
altitudes from 6 to 500~m and camera down-tilt angles from
$5^\circ$ to $90^\circ$, providing both oblique and near-nadir
observations under varied scene structures. It provides georeferenced aerial and satellite reference maps. We use the aerial reference maps, whose corresponding DSMs have spatial resolutions ranging from 0.15 to 1.98~m, and follow the reference-map rotation and cropping protocol of \cite{ye2026deco} to construct the UAV-image--reference-map pairs.

\textbf{OrthoLoC} \cite{dhaouadi2025ortholoc} contains 16,427 UAV samples from 51 locations, including 10,923
same-domain and 5,504 cross-domain samples. Its flight altitudes
range from 23 to 201~m, and most observations have down-tilt
angles between $60^\circ$ and $90^\circ$. The dataset provides georeferenced orthophotos and corresponding DSMs with spatial resolutions ranging from 0.04 to 0.73~m. The same-domain reference maps are generated
from aerial photogrammetry, whereas the cross-domain setting
introduces reference data from different sources and acquisition
conditions. Following \cite{ye2026deco}, we use all provided samples and restrict the yaw difference between UAV image and reference map to
$45^\circ$. These two datasets provide complementary scene structures, viewing
conditions, and reference-map settings for real-world evaluation.

\subsubsection{Comparison with Existing Methods}
\label{sec:real_world_comparison}

Table~\ref{tab:real_world_main} reports the real-world localization results
using RoMa correspondences. DSM-PnP provides a DSM-based reference and is therefore excluded when ranking the
DSM-free methods. Among the OPG-based DSM-free methods, LDVP consistently outperforms AH and
OPE on both datasets. The improvement is particularly clear at $A@1$,
supporting the effectiveness of the proposed direct metric pose formulation. The proposed SAC further improves LDVP on both datasets, with particularly clear gains at
$A@3$ and $A@5$. On AnyVisLoc, SAC improves $A@3$ by 6.10 percentage points.
On OrthoLoC, SAC also improves $A@3$ and $A@5$, although $A@1$ decreases
slightly. This indicates that the Planar-PnP branch mainly helps reduce larger localization errors for near-planar queries, while its benefit can be less evident for queries that LDVP already localizes accurately.

\begin{figure*}[!h]
    \centering
    \includegraphics[width=\textwidth]
    {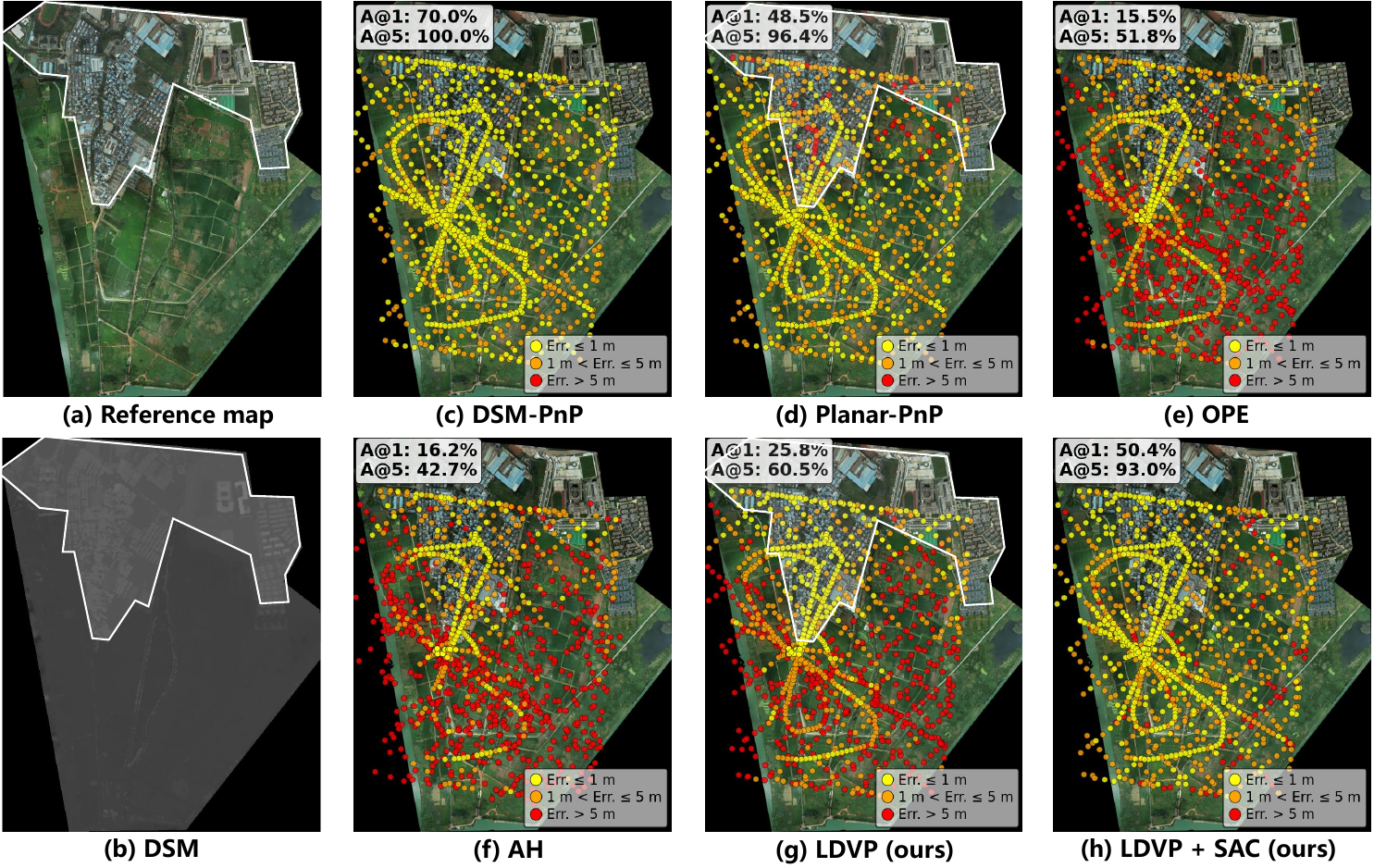}
    \caption{
    Scene-level localization results for Case 1 (AnyVisLoc Scene 07), a
    mixed urban--farmland environment. The upper region of the reference
    map (white polygon) contains pronounced urban height variations, whereas most of the
    remaining area is relatively flat farmland. The panels show the reference
    map, DSM, and localization results of DSM-PnP, Planar-PnP, OPE, AH, LDVP,
    and LDVP+SAC. Localization errors are color-coded using the $1$~m and
    $5$~m thresholds.
    }
    \label{fig:scene07_localization}
\end{figure*}

\begin{figure*}[!h]
    \centering
    \begin{minipage}[h]{0.495\textwidth}
        \centering
        \includegraphics[width=\linewidth]
        {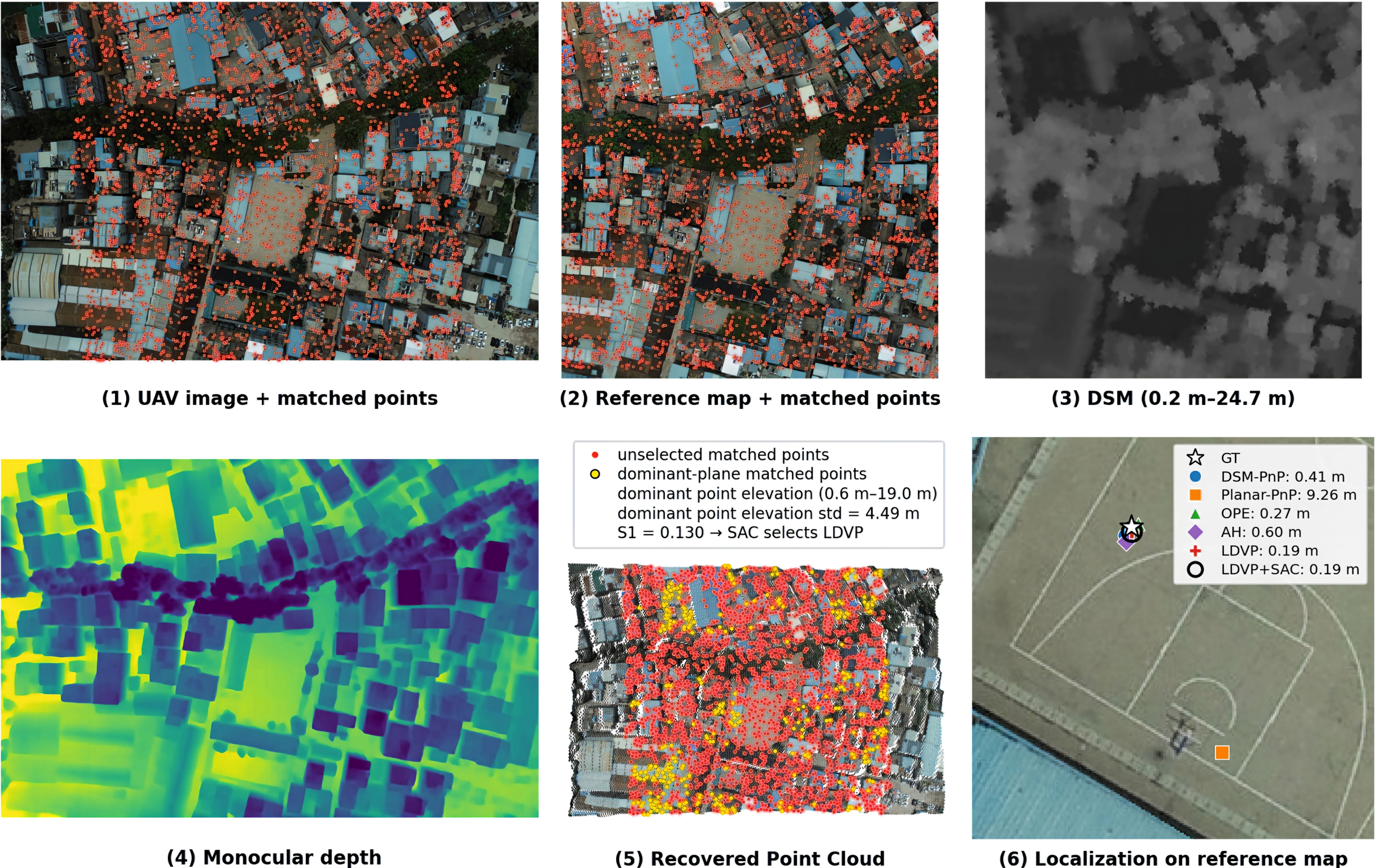}
        \subcaption{Urban query with $S_1=0.130$.}
        \label{fig:scene07_urban_sample}
    \end{minipage}
    \hfill
    \begin{minipage}[h]{0.495\textwidth}
        \centering
        \includegraphics[width=\linewidth]
        {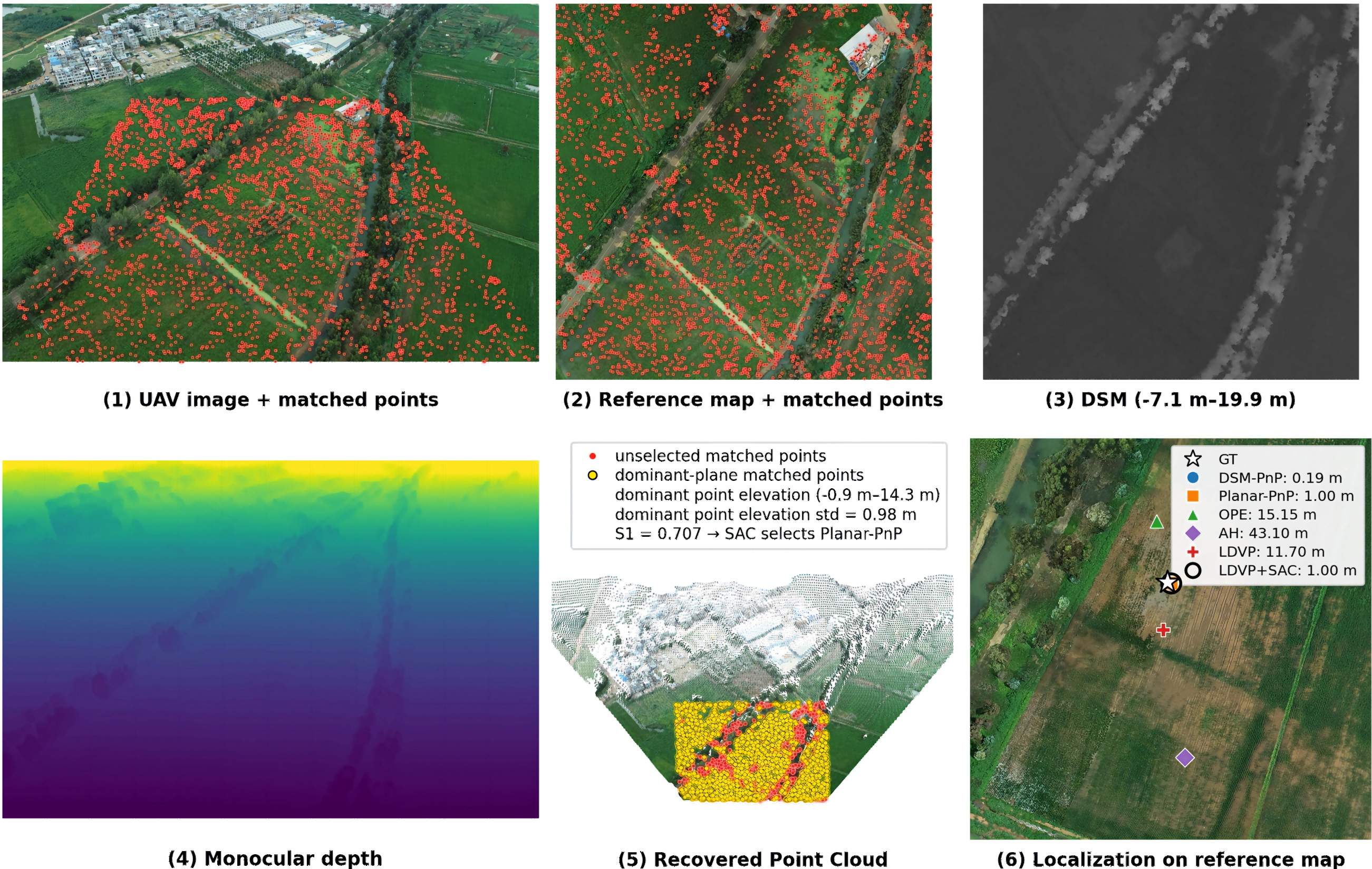}
        \subcaption{Farmland query with $S_1=0.707$.}
        \label{fig:scene07_farmland_sample}
    \end{minipage}
    \caption{
    Representative SAC decisions in Case 1 (AnyVisLoc Scene 07).
    Each example visualizes the UAV--reference matched points, reference DSM,
    monocular depth, recovered 3D structure and dominant-plane support,
    and the resulting localization estimates. DSM elevations are shown only
    for retrospective analysis and are not used by SAC.
    }
    \label{fig:scene07_sac_examples}
\end{figure*}
\subsubsection{Analysis of Scene Structure}
\label{sec:structure_analysis}
To understand how SAC benefits from scene structure, we analyze how scene
relief affects the relative performance of LDVP and Planar-PnP and whether
SAC adapts its solver selection accordingly in
Fig.~\ref{fig:real_dsm_structure}. As shown in Fig.~\ref{fig:real_dsm_structure}(a), AnyVisLoc contains substantially larger scene relief than OrthoLoC. This difference may partly explain the larger gains of SAC on AnyVisLoc in Table~\ref{tab:real_world_main}, where the greater variation in scene structure provides more opportunities to benefit from the complementary strengths of LDVP and Planar-PnP. Fig.~\ref{fig:real_dsm_structure}(b) further reveals a clear complementary trend between LDVP and Planar-PnP: LDVP is more likely to yield lower error as scene relief increases, whereas Planar-PnP is favored in low-relief regions. This trend is consistent with the scene-dependent behavior predicted by the observability analysis and observed in the synthetic experiment. The transition is smoother than in the controlled synthetic experiment, reflecting the more complex scene structures and correspondence distributions in real-world data.

\begin{figure*}[t]
    \centering
    \includegraphics[width=\textwidth]
    {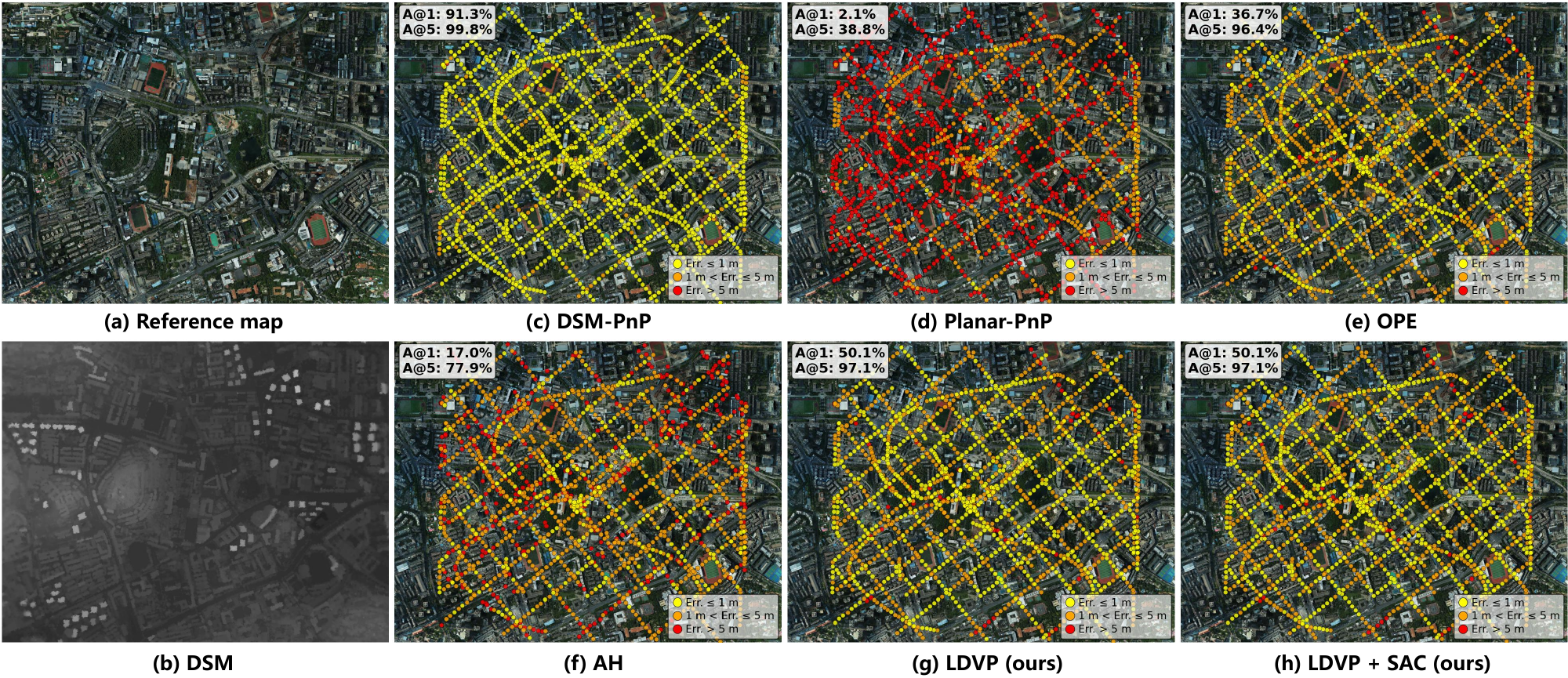}
    \caption{
    Scene-level localization results for Case 2 (AnyVisLoc Scene 03), a
    dense urban environment with strong multi-level elevation structure.
    Planar-PnP degrades substantially in this regime, whereas LDVP and
    LDVP+SAC remain more stable.
    }
    \label{fig:scene03_localization}
\end{figure*}

\begin{figure}[t]
    \centering
    \includegraphics[width=\linewidth]
    {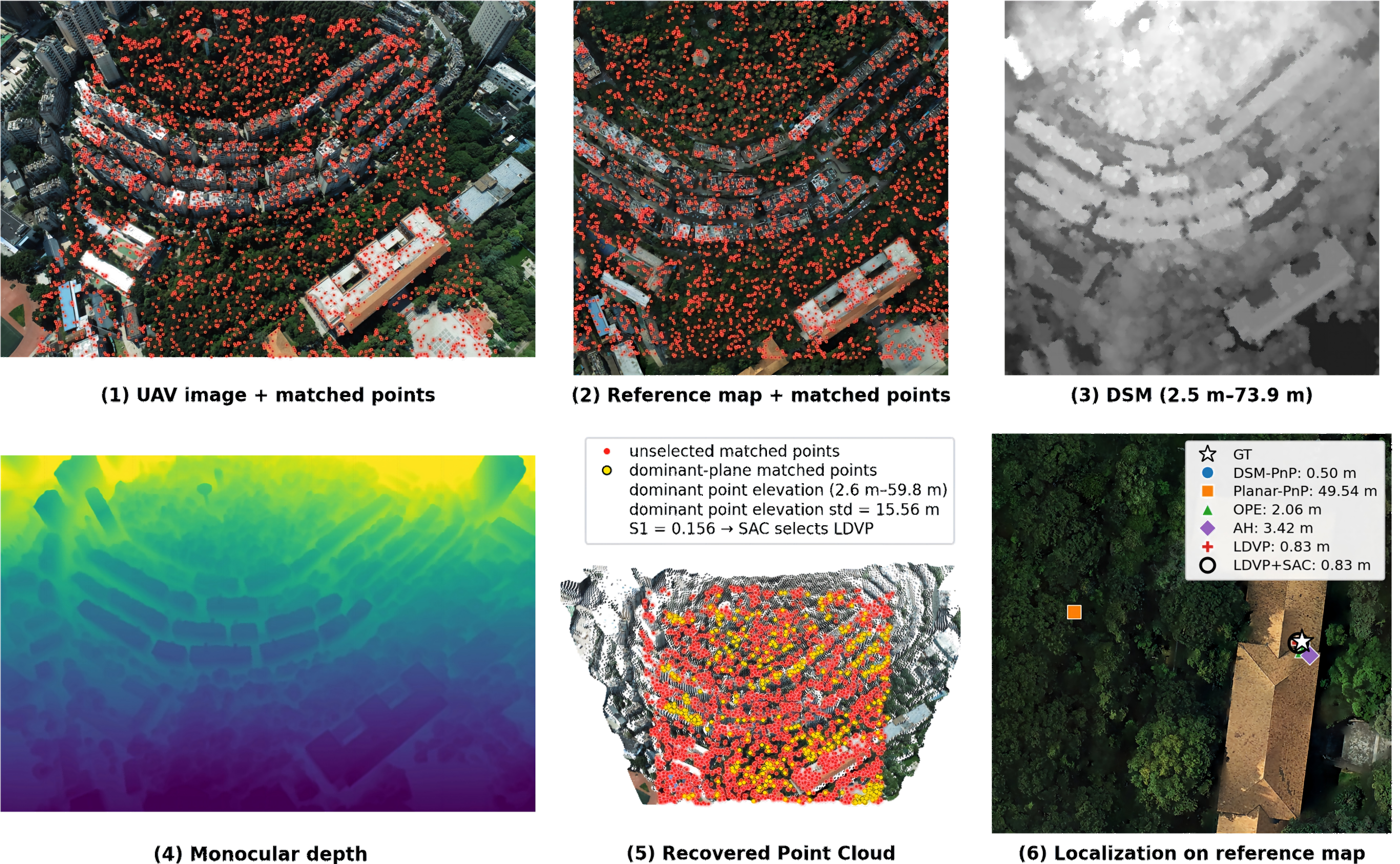}
    \caption{
    Representative query from Case 2 (AnyVisLoc Scene 03).
    The matched points span strongly varying
elevations, with a dominant-plane support ratio
of $S_1=0.156$; SAC therefore selects LDVP.
    }
    \label{fig:scene03_sac_example}
\end{figure}

Fig.~\ref{fig:real_dsm_structure}(c) shows a corresponding change in the SAC
decisions. LDVP is selected increasingly often as scene relief increases,
whereas Planar-PnP is favored at small relief. Finally,
Fig.~\ref{fig:real_dsm_structure}(d) shows that the agreement between the
SAC-selected branch and the lower-error solver is higher at small and large
relief but decreases in the intermediate $2$--$10$~m range. This is
consistent with Fig.~\ref{fig:real_dsm_structure}(b), where the relative
advantage of LDVP and Planar-PnP is less distinct in this intermediate range.
Overall, the real-world results support the structure-dependent
complementarity identified in the synthetic experiment and show that SAC
largely follows this behavior when selecting between the two solvers.

\subsubsection{Qualitative Analysis of SAC}
\label{sec:sac_qualitative}

We further visualize two representative AnyVisLoc scenes with different
elevation structures and representative query-level examples from OrthoLoC
to illustrate the complementary behavior of LDVP and Planar-PnP and the
corresponding solver adaptation of SAC. Reference DSMs, when available, are
shown only for retrospective geometric interpretation and are not used by
SAC during localization.
\begin{figure*}[!t]
    \centering
    \begin{minipage}[t]{0.495\textwidth}
        \centering
        \includegraphics[width=\linewidth]
        {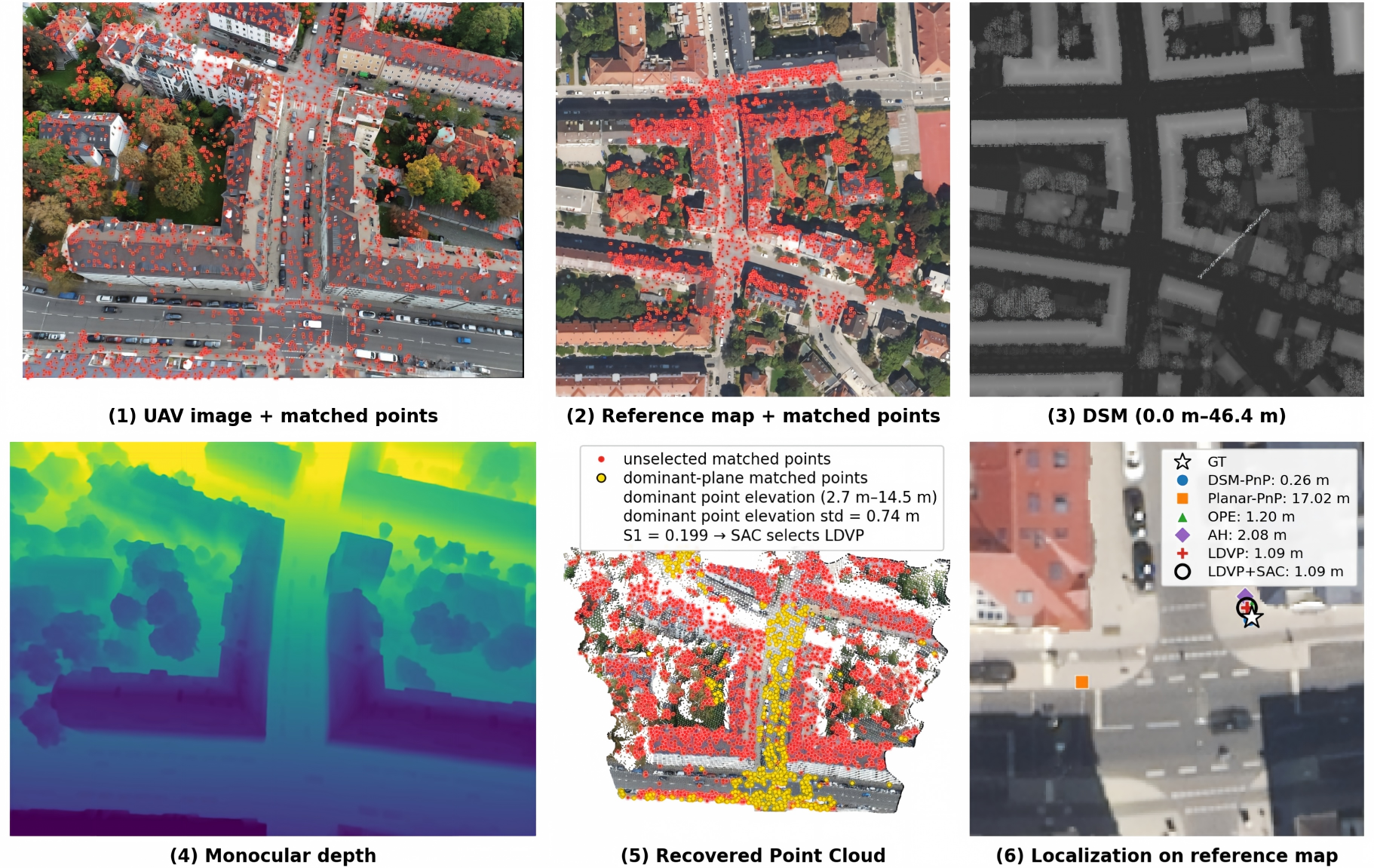}
        \subcaption{LDVP selected, $S_1=0.199$.}
        \label{fig:ortholoc_ldvp_1}
    \end{minipage}
    \hfill
    \begin{minipage}[t]{0.495\textwidth}
        \centering
        \includegraphics[width=\linewidth]
        {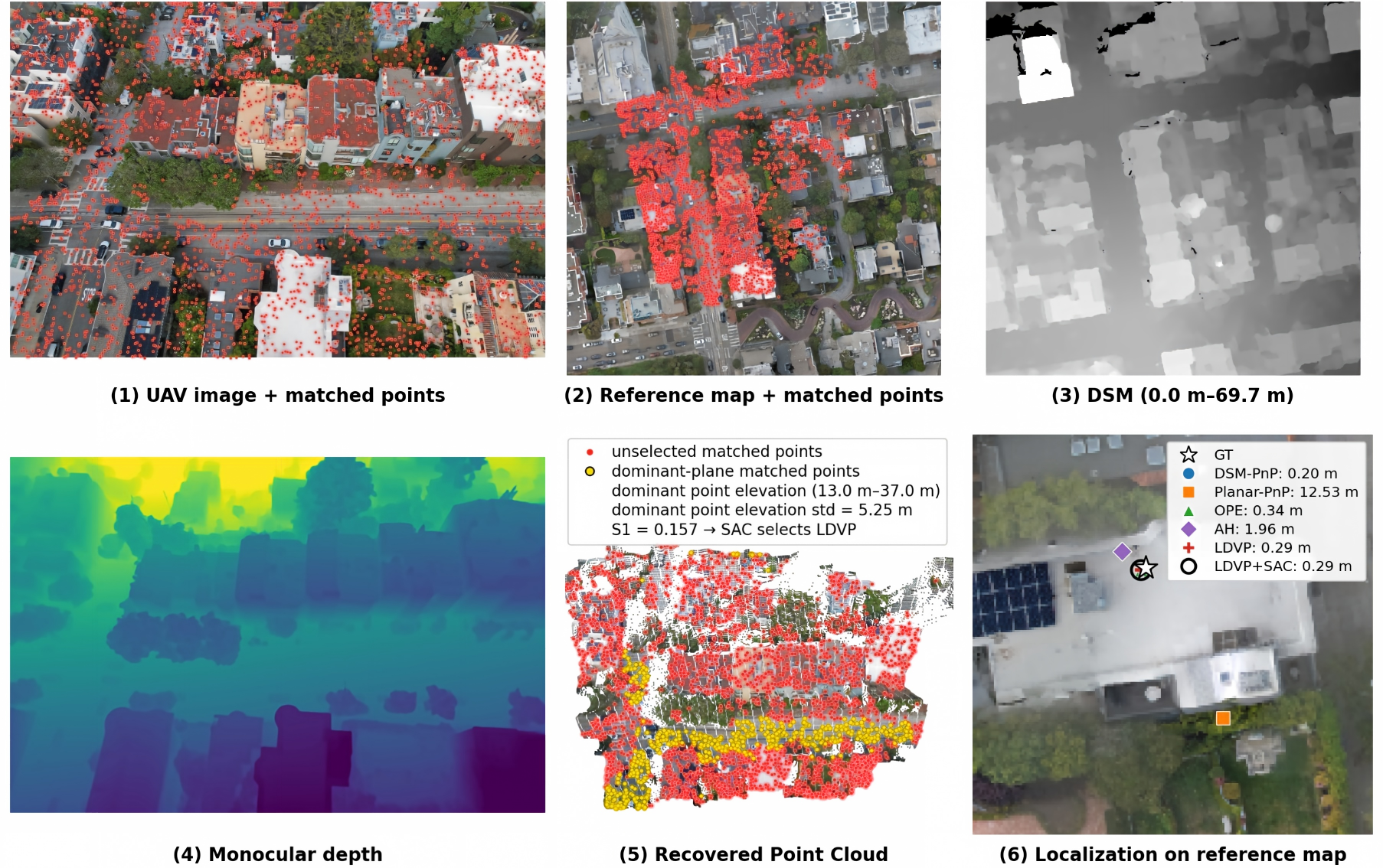}
        \subcaption{LDVP selected, $S_1=0.157$.}
        \label{fig:ortholoc_ldvp_2}
    \end{minipage}

    \vspace{1.5mm}

    \begin{minipage}[t]{0.495\textwidth}
        \centering
        \includegraphics[width=\linewidth]
        {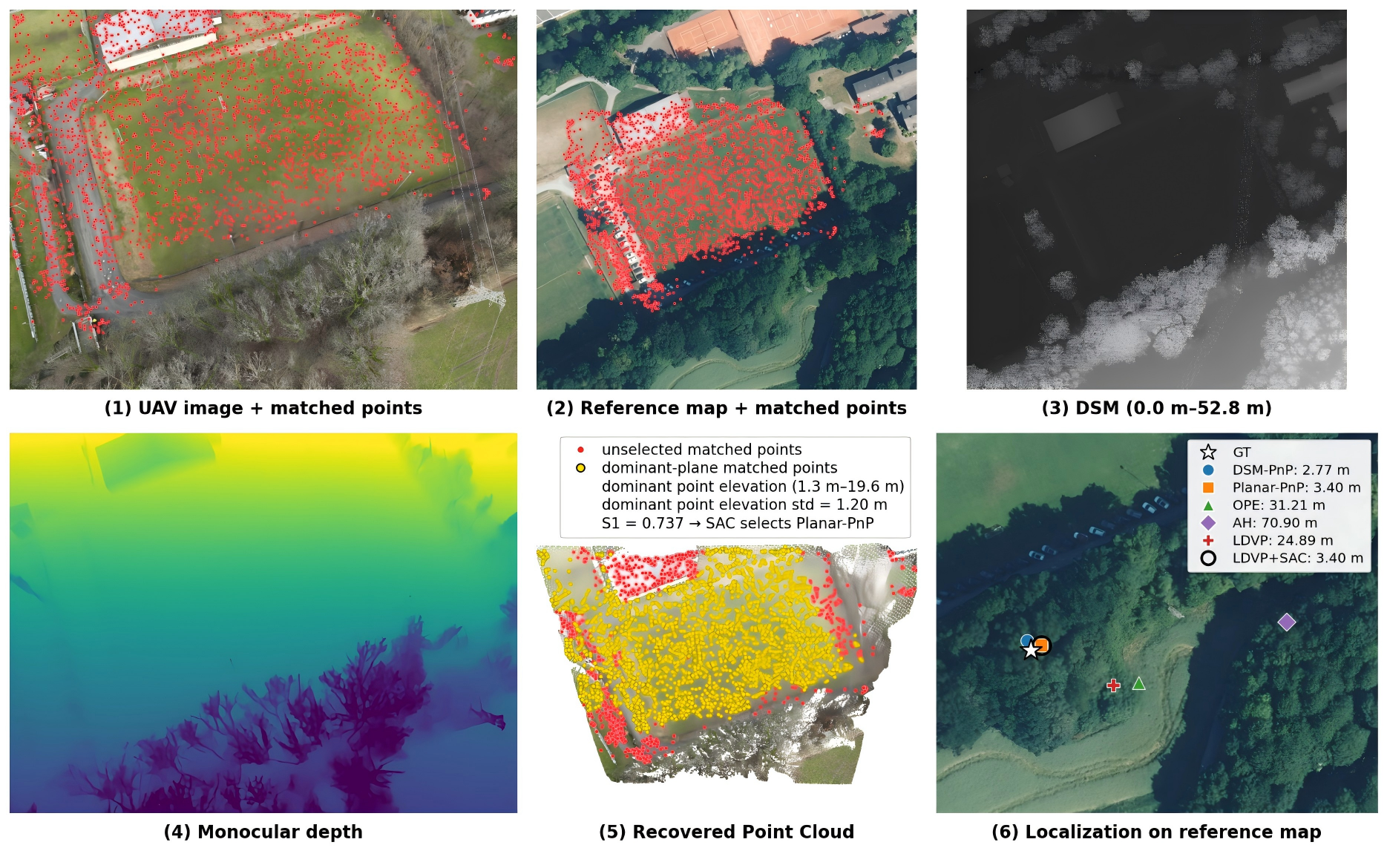}
        \subcaption{Planar-PnP selected, $S_1=0.737$.}
        \label{fig:ortholoc_planar_1}
    \end{minipage}
    \hfill
    \begin{minipage}[t]{0.495\textwidth}
        \centering
        \includegraphics[width=\linewidth]
        {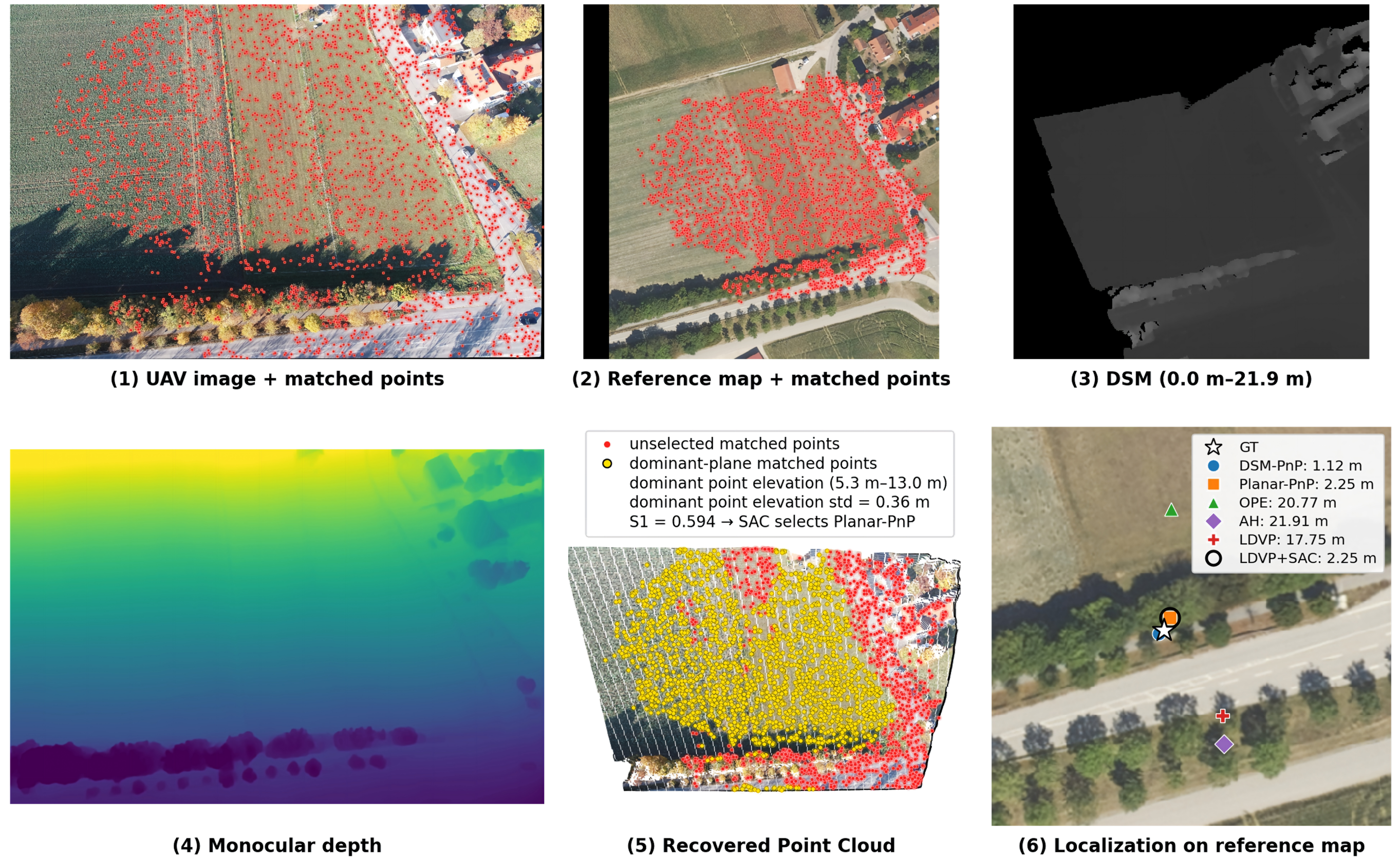}
        \subcaption{Planar-PnP selected, $S_1=0.594$.}
        \label{fig:ortholoc_planar_2}
    \end{minipage}

    \caption{
    Representative SAC decisions on the OrthoLoC dataset. The top row shows
    two queries with weak dominant-plane support, for which SAC selects LDVP,
    while the bottom row shows two queries with strong dominant-plane support,
    for which SAC switches to Planar-PnP. Each example visualizes the UAV--reference matched points, reference DSM,
    monocular depth, recovered 3D structure and dominant-plane support,
    and the resulting localization estimates.
    }
    \label{fig:ortholoc_sac_examples}
\end{figure*}

\noindent\textbf{AnyVisLoc Case 1: mixed urban--farmland scene.}
Fig.~\ref{fig:scene07_localization} shows clear spatial complementarity
between LDVP and Planar-PnP within the same scene. LDVP performs reliably in
the urban region enclosed by the white polygon, where the terrain and
buildings exhibit substantial elevation variation, but produces more large
errors over the relatively flat farmland. Planar-PnP shows the opposite
behavior: most of its errors above $5$~m are concentrated around the urban
region, whereas it achieves higher accuracy over the farmland, where the
matched scene is largely dominated by planar structure. This spatial pattern
is consistent with the scene-dependent behavior identified in the
observability analysis and synthetic experiments. SAC adapts to this variation
by favoring LDVP in the urban region and Planar-PnP over much of the farmland,
thereby improving the overall localization performance. The selection is not
always correct, however. Some queries in the lower-right farmland region are
still routed to LDVP and result in errors above $5$~m, which explains the
approximately $3.4$ percentage-point decrease in $A@5$ compared with
Planar-PnP. These cases suggest that monocular depth may not always provide
sufficiently reliable relative structure for identifying planar queries in
relatively homogeneous rural regions.

The representative queries in Fig.~\ref{fig:scene07_sac_examples} illustrate
this behavior more explicitly. In Fig.~\ref{fig:scene07_sac_examples}(a), the
matched points are distributed over building surfaces at different
elevations, and only $S_1=0.130$ of the valid matched points support the
dominant plane. SAC therefore retains LDVP, whereas Planar-PnP produces a
localization error of $9.26$~m. In the farmland query shown in Fig.~\ref{fig:scene07_sac_examples}(b), most matched points lie
on terrain with similar elevations and the dominant-plane support increases
to $S_1=0.707$. SAC consequently switches to Planar-PnP, reducing the
localization error from $11.7$~m with LDVP to $1.00$~m. These examples
directly illustrate how the scene structure underlying the matched points
leads to different favorable solvers and how SAC exploits this
complementarity at the query level.


\noindent\textbf{AnyVisLoc Case 2: dense urban scene.}
Fig.~\ref{fig:scene03_localization} shows a dense urban scene with strong elevation variation. Planar-PnP produces large errors for many queries because the matched points are distributed over surfaces at substantially different elevations, violating the planar assumption. In contrast, the OPG-based methods remain substantially more accurate in this scene, with LDVP achieving the best overall performance. Since all queries have $S_1<0.5$, SAC consistently retains LDVP throughout the scene. The representative query in Fig.~\ref{fig:scene03_sac_example} further illustrates this behavior: the matched points span elevations from \(2.5\)~m to \(73.9\)~m, while the dominant-plane support is only $S_1=0.156$. Under this strongly non-planar structure, LDVP achieves a localization error of \(0.83\)~m, whereas Planar-PnP fails to provide accurate localization. SAC accordingly selects LDVP, consistent with the substantial elevation variation of the matched scene.


\noindent\textbf{OrthoLoC.}
Since OrthoLoC does not provide the continuous scene-level reference maps
used for the AnyVisLoc visualization, we report query-level examples only. Fig.~\ref{fig:ortholoc_sac_examples} shows representative query-level
decisions on OrthoLoC. SAC retains LDVP for queries with weak dominant-plane
support and switches to Planar-PnP when the dominant-plane support exceeds
the threshold, showing the same structure-adaptive behavior as on AnyVisLoc.
\begin{table}[h]
    \centering
    \caption{Sensitivity to the SAC consensus threshold using RoMa.
    $\Delta A@\tau$ denotes the percentage-point change relative to LDVP;
    SelP denotes the percentage of queries assigned to Planar-PnP.}
    \label{tab:sac_sensitivity}
    \setlength{\tabcolsep}{4.0pt}
    \renewcommand{\arraystretch}{1.08}
    \resizebox{\columnwidth}{!}{
    \begin{tabular}{lrrrr}
        \toprule
        Dataset / $\tau_S$ & SelP (\%) & $\Delta A@1$ & $\Delta A@3$ & $\Delta A@5$ \\
        \midrule
        AnyVisLoc / 0.30 & 33.2 & +0.74 & +6.72 & +6.99 \\
        AnyVisLoc / 0.40 & 23.3 & +2.61 & +7.21 & +6.85 \\
       AnyVisLoc / \textbf{0.50} & 14.9 & +2.76 & +6.10 & +5.49 \\
        AnyVisLoc / 0.60 & 8.3 & +1.84 & +3.84 & +3.50 \\
        \midrule
        OrthoLoC / 0.30 & 57.02 & -4.87 & +4.72 & +8.59 \\
        OrthoLoC / 0.40 & 33.47 & -1.91 & +4.52 & +6.71 \\
        OrthoLoC / \textbf{0.50} & 17.06 & -0.55 & +2.77 & +4.10 \\
        OrthoLoC / 0.60 & 7.10 & -0.09 & +1.25 & +1.85 \\
        \bottomrule
    \end{tabular}}
\end{table}

\begin{table*}[!t]
    \centering
    \caption{Localization results with different matching methods. Best and
    second-best DSM-free results are shown in \textbf{bold} and
    \underline{underline}, respectively. DSM-PnP is excluded from the ranking.}
    \label{tab:matching_ablation}
    \setlength{\tabcolsep}{5pt}
    \renewcommand{\arraystretch}{1.08}
    \resizebox{\textwidth}{!}{
    \begin{tabular}{llrrrrrrrr}
        \toprule
        \multirow{2}{*}{Matching method} & \multirow{2}{*}{Localization method}
        & \multicolumn{4}{c}{AnyVisLoc} & \multicolumn{4}{c}{OrthoLoC} \\
        \cmidrule(lr){3-6}\cmidrule(lr){7-10}
        & & $A@1$ (\%) & $A@3$ (\%) & $A@5$ (\%) & Med. (m)
          & $A@1$ (\%) & $A@3$ (\%) & $A@5$ (\%) & Med. (m) \\
        \midrule
        \multirow{6}{*}{SuperPoint+LightGlue}
        & \cellcolor{gray!15}DSM-PnP
        & \cellcolor{gray!15}41.23 & \cellcolor{gray!15}84.55
        & \cellcolor{gray!15}91.92 & \cellcolor{gray!15}1.20
        & \cellcolor{gray!15}49.45 & \cellcolor{gray!15}61.25
        & \cellcolor{gray!15}63.16 & \cellcolor{gray!15}0.83 \\
        & Planar-PnP & 7.98 & 33.20 & 49.99 & 4.93
                     & 9.21 & 34.92 & 47.82 & \underline{4.52} \\
        & AH & 8.07 & 23.87 & 33.28 & 11.37
             & 10.77 & 23.50 & 29.70 & 23.50 \\
        & OPE & 6.22 & 21.83 & 32.18 & 9.91
              & 13.45 & 29.93 & 39.19 & 8.02 \\
        & LDVP (ours) & \underline{17.77} & \underline{38.65}
                      & \underline{50.73} & \underline{4.77}
                      & \underline{23.25} & \underline{40.54}
                      & \underline{47.95} & 4.76 \\
        & LDVP+SAC (ours) & \textbf{20.35} & \textbf{47.11}
                          & \textbf{60.69} & \textbf{3.29}
                          & \textbf{23.42} & \textbf{44.94}
                          & \textbf{53.30} & \textbf{3.28} \\
        \midrule
        \multirow{6}{*}{DISK+LightGlue}
        & \cellcolor{gray!15}DSM-PnP
        & \cellcolor{gray!15}57.18 & \cellcolor{gray!15}88.78
        & \cellcolor{gray!15}91.46 & \cellcolor{gray!15}0.84
        & \cellcolor{gray!15}42.54 & \cellcolor{gray!15}53.40
        & \cellcolor{gray!15}55.83 & \cellcolor{gray!15}0.71 \\
        & Planar-PnP & 9.33 & 36.37 & 51.55 & 4.55
                     & 5.59 & 26.21 & 37.88 & 4.50 \\
        & AH & 9.44 & 24.53 & 32.78 & 13.62
             & 5.97 & 15.13 & 20.05 & 28.32 \\
        & OPE & 11.16 & 31.03 & 42.07 & 6.79
              & 12.06 & 27.02 & 34.72 & 8.49 \\
        & LDVP (ours) & \underline{25.05} & \underline{45.60}
                      & \underline{56.13} & \underline{3.54}
                      & \underline{19.12} & \underline{33.46}
                      & \underline{39.97} & \underline{3.90} \\
        & LDVP+SAC (ours) & \textbf{28.59} & \textbf{57.04}
                          & \textbf{68.79} & \textbf{2.22}
                          & \textbf{19.46} & \textbf{37.35}
                          & \textbf{44.37} & \textbf{2.83} \\
        \midrule
        \multirow{6}{*}{ALIKED+LightGlue}
        & \cellcolor{gray!15}DSM-PnP
        & \cellcolor{gray!15}66.59 & \cellcolor{gray!15}92.25
        & \cellcolor{gray!15}94.03 & \cellcolor{gray!15}0.72
        & \cellcolor{gray!15}47.87 & \cellcolor{gray!15}60.19
        & \cellcolor{gray!15}62.45 & \cellcolor{gray!15}0.86 \\
        & Planar-PnP & 9.63 & 38.30 & 54.13 & 4.26
                     & 7.82 & 31.77 & 44.88 & \underline{5.03} \\
        & AH & 12.44 & 31.59 & 41.66 & 7.66
             & 9.98 & 23.21 & 29.32 & 27.28 \\
        & OPE & 14.00 & 36.56 & 49.44 & 5.10
              & 14.70 & 31.63 & 40.07 & 8.19 \\
        & LDVP (ours) & \underline{25.24} & \underline{49.13}
                      & \underline{60.95} & \underline{3.09}
                      & \underline{23.64} & \underline{39.06}
                      & \underline{46.38} & 5.19 \\
        & LDVP+SAC (ours) & \textbf{28.82} & \textbf{58.91}
                          & \textbf{71.26} & \textbf{2.21}
                          & \textbf{23.94} & \textbf{42.35}
                          & \textbf{49.97} & \textbf{3.75} \\
        \bottomrule
    \end{tabular}}
\end{table*}

\subsection{Ablation Studies}

\subsubsection{SAC Threshold}
\label{sec:sac_sensitivity}

We evaluate the sensitivity of SAC to the consensus threshold $\tau_S$ using
RoMa correspondences. Table~\ref{tab:sac_sensitivity} reports the Planar-PnP
selection rate and the change in localization accuracy relative to LDVP. We vary $\tau_S$ from 0.30 to 0.60. The threshold is varied only for this
analysis; all main experiments use $\tau_S=0.5$. Lowering $\tau_S$ increases the use of Planar-PnP and generally produces
larger gains at $A@3$ and $A@5$, but can reduce $A@1$ on OrthoLoC.
Increasing $\tau_S$ reduces the Planar-PnP selection rate and gradually
brings SAC closer to LDVP. We use $\tau_S=0.5$ in all main experiments as a
fixed majority threshold across both datasets.

\subsubsection{Image Matching Methods}
\label{sec:matching_ablation}

We further examine the influence of the image matching method by replacing
the default dense matcher RoMa with three representative sparse matching
pipelines based on SuperPoint \cite{detone18superpoint}, DISK \cite{DISK},
and ALIKED \cite{zhao2023aliked}, respectively paired with LightGlue
\cite{lightglue}. As shown in Table~\ref{tab:matching_ablation}, LDVP consistently achieves higher \(A@1\) than OPE and AH across both datasets with all three matching pipelines, confirming that its advantage is not specific to RoMa. SAC further improves LDVP at $A@3$ and $A@5$ and reduces the median error in every case.

\subsubsection{Monocular Depth Models}
\label{sec:depth_ablation}

\begin{table}[t]
    \centering
    \caption{Ablation of the monocular depth model used by SAC with RoMa
    correspondences and $\tau_S=0.5$. $A@1$, $A@3$, and $A@5$
    are reported in percentages (\%), and Med. in meters (m).}
    \label{tab:depth_ablation}
    \setlength{\tabcolsep}{4.0pt}
    \renewcommand{\arraystretch}{1.08}
    \resizebox{\columnwidth}{!}{
    \begin{tabular}{lrrrrrrrr}
        \toprule
        \multirow{2}{*}{Method}
        & \multicolumn{4}{c}{AnyVisLoc}
        & \multicolumn{4}{c}{OrthoLoC} \\
        \cmidrule(lr){2-5}\cmidrule(lr){6-9}
        & $A@1$ & $A@3$ & $A@5$ & Med.
        & $A@1$ & $A@3$ & $A@5$ & Med. \\
        \midrule
        LDVP
        & 48.59 & 75.41 & 82.87 & 1.05
        & 39.89 & 59.28 & 66.23 & 1.69 \\
        \midrule
        LDVP+SAC (MoGe2-Aerial)
        & 51.35 & 81.51 & 88.36 & 0.97
        & 39.34 & 62.05 & 70.33 & 1.66 \\
        LDVP+SAC (MoGe2)
        & 51.19 & 80.92 & 87.87 & 0.97
        & 39.72 & 61.62 & 69.57 & 1.64 \\
        LDVP+SAC (Depth Anything 3)
        & 51.25 & 81.64 & 88.50 & 0.97
        & 38.53 & 62.49 & 71.34 & 1.70 \\
        \bottomrule
    \end{tabular}}
\end{table}

We further evaluate the influence of the monocular depth model used by SAC.
In addition to MoGe2-Aerial \cite{moge2aerial}, which adapts MoGe2 \cite{moge2} to aerial
imagery, we test the original MoGe2 and Depth Anything 3 \cite{DA3}. As shown in
Table~\ref{tab:depth_ablation}, SAC consistently improves LDVP at $A@3$ and
$A@5$ with all three depth models on both datasets. The differences among
the three models are relatively small, despite their different depth
predictions. This indicates that SAC does not depend strongly on a specific
monocular depth model, consistent with its use of predicted depth only to
infer the relative structure of the matched points rather than as metric
3D input for pose estimation.

%% file: sec/6_conclusion.tex
\section{Conclusion}

This work presented a DSM-free UAV localization
framework for direct metric pose estimation without
requiring external elevation information.
We proposed LDVP to directly recover the metric
horizontal camera position under OPG by analytically
eliminating latent ray depths.
Our local observability analysis revealed how scene
elevation variation affects the pose constraints of OPG,
motivating SAC to adaptively select between LDVP and
Planar-PnP using relative scene structure inferred
from monocular depth estimation.
Experiments demonstrated that LDVP outperforms existing
OPG-based DSM-free methods, while SAC further improves
robustness across diverse scenes and substantially
narrows the gap to DSM-based localization.

The demonstrated accuracy of geometric pose estimation without DSM information, together with the ability of monocular depth to capture scene structure, suggests the potential for end-to-end DSM-free localization. Future work could integrate scene-structure understanding and pose estimation into a unified learned framework, with the geometric insights developed here providing a basis for its design.

%% file: sec/7_appendix_revised_v3.tex
{\appendices
\section{Detailed Relation among LDVP, OPE, and AH}
\label{supp:ope_ah_relation}

This section derives Eq.~\eqref{eq:ope_ah_ldvp_relation_main} in the main
paper. The derivation shows that LDVP, OPE \cite{Larsson_2021_ICCV}, and AH
\cite{Li2026AH} encode the same noiseless orthographic--perspective incidence
constraint, although they use different residual parameterizations.

For the $i$-th correspondence, define the homogeneous orthographic map point
\begin{equation}
    \overline{\mathbf{m}}_i
    =
    \begin{bmatrix}
        X_i & Y_i & 1
    \end{bmatrix}^{\top},
\end{equation}
and write the world-to-camera rotation as
\begin{equation}
    \mathbf{R}
    =
    \begin{bmatrix}
        \mathbf{a}_1 &
        \mathbf{a}_2 &
        \mathbf{a}_3
    \end{bmatrix}
    \in \mathrm{SO}(3).
    \label{eq:supp_rotation_columns}
\end{equation}
Since
\begin{equation}
    \mathbf{R}^{\top}\mathbf{r}_i
    =
    \begin{bmatrix}
        \mathbf{a}_1^{\top}\mathbf{r}_i\\
        \mathbf{a}_2^{\top}\mathbf{r}_i\\
        \mathbf{a}_3^{\top}\mathbf{r}_i
    \end{bmatrix},
\end{equation}
the horizontal ray components in Eq.~\eqref{eq:world_bearing} satisfy
\begin{equation}
    q_{ix}=\mathbf{a}_1^{\top}\mathbf{r}_i,
    \qquad
    q_{iy}=\mathbf{a}_2^{\top}\mathbf{r}_i.
    \label{eq:supp_q_components}
\end{equation}

Under the pose convention of the main paper, the OPE matrix is
\begin{equation}
    \mathbf{E}_{\mathrm{OP}}
    =
    \begin{bmatrix}
        -\mathbf{a}_2^{\top}\\
        \mathbf{a}_1^{\top}\\
        C_x\mathbf{a}_2^{\top}-C_y\mathbf{a}_1^{\top}
    \end{bmatrix}.
    \label{eq:supp_ope_matrix}
\end{equation}
Its incidence residual expands to
\begin{align}
    \overline{\mathbf{m}}_i^{\top}
    \mathbf{E}_{\mathrm{OP}}\mathbf{r}_i
    &=
    -q_{iy}X_i+q_{ix}Y_i+C_xq_{iy}-C_yq_{ix}
    \nonumber\\
    &=
    -q_{iy}(X_i-C_x)+q_{ix}(Y_i-C_y).
    \label{eq:supp_ope_expansion}
\end{align}

For AH, let
\begin{equation}
    \mathbf{t}
    =
    -\mathbf{R}\mathbf{C}
    =
    -C_x\mathbf{a}_1-C_y\mathbf{a}_2-C_z\mathbf{a}_3.
    \label{eq:supp_translation}
\end{equation}
The corresponding matrix is
\begin{equation}
    \mathbf{E}_{\mathrm{AH}}
    =
    [\mathbf{a}_3]_{\times}
    \begin{bmatrix}
        \mathbf{a}_1 & \mathbf{a}_2 & \mathbf{t}
    \end{bmatrix}.
    \label{eq:supp_ah_matrix}
\end{equation}
Because the columns of $\mathbf{R}$ form a right-handed orthonormal basis,
\begin{equation}
    \mathbf{a}_3\times\mathbf{a}_1=\mathbf{a}_2,
    \qquad
    \mathbf{a}_3\times\mathbf{a}_2=-\mathbf{a}_1,
\end{equation}
and therefore
\begin{equation}
    [\mathbf{a}_3]_{\times}\mathbf{t}
    =
    -C_x\mathbf{a}_2+C_y\mathbf{a}_1.
\end{equation}
It follows that
\begin{equation}
    \mathbf{E}_{\mathrm{AH}}
    =
    \begin{bmatrix}
        \mathbf{a}_2 &
        -\mathbf{a}_1 &
        -C_x\mathbf{a}_2+C_y\mathbf{a}_1
    \end{bmatrix}
    =
    -\mathbf{E}_{\mathrm{OP}}^{\top}.
    \label{eq:supp_ah_ope_relation}
\end{equation}

The LDVP residual in Eq.~\eqref{eq:ldvp_residual} is
\begin{equation}
    e_i
    =
    \frac{
        -q_{iy}(X_i-C_x)+q_{ix}(Y_i-C_y)
    }{
        \|\mathbf{q}_i\|_2
    },
    \qquad
    \|\mathbf{q}_i\|_2>0.
    \label{eq:supp_ldvp_residual}
\end{equation}
Combining Eqs.~\eqref{eq:supp_ope_expansion},
\eqref{eq:supp_ah_ope_relation}, and
\eqref{eq:supp_ldvp_residual} gives
\begin{equation}
    \mathbf{r}_i^{\top}
    \mathbf{E}_{\mathrm{AH}}
    \overline{\mathbf{m}}_i
    =
    -
    \overline{\mathbf{m}}_i^{\top}
    \mathbf{E}_{\mathrm{OP}}
    \mathbf{r}_i
    =
    -
    \|\mathbf{q}_i\|_2 e_i.
    \label{eq:supp_threeway_relation}
\end{equation}
Hence all three residuals vanish for the same noiseless
UAV--reference correspondence geometry. Their distinction lies in the
estimation parameterization: OPE and AH use an intermediate matrix, whereas
LDVP evaluates the incidence error directly in pose space.

\section{Variable Projection in LDVP}
\label{supp:variable_projection}

The main paper gives the closed-form latent scale in
Eq.~\eqref{eq:lambda_ldvp}. Here we show why its elimination produces the
metric point-to-line residual exactly.

For fixed $\mathbf{R}$ and $\mathbf{c}$, let
\begin{equation}
    \mathbf{d}_i=\mathbf{m}_i-\mathbf{c},
    \qquad
    f_i(\lambda_i)
    =
    \|\mathbf{d}_i-\lambda_i\mathbf{q}_i\|_2^2.
\end{equation}
Assuming $\|\mathbf{q}_i\|_2>0$, minimizing this quadratic gives
\begin{equation}
    \lambda_i^{\ast}
    =
    \frac{\mathbf{q}_i^{\top}\mathbf{d}_i}
         {\mathbf{q}_i^{\top}\mathbf{q}_i}.
    \label{eq:supp_lambda_closed_form}
\end{equation}
Substitution yields
\begin{equation}
\begin{split}
    f_i(\lambda_i^{\ast})
    &=
    \mathbf{d}_i^{\top}
    \left(
        \mathbf{I}_2
        -
        \frac{\mathbf{q}_i\mathbf{q}_i^{\top}}
             {\mathbf{q}_i^{\top}\mathbf{q}_i}
    \right)
    \mathbf{d}_i .
\end{split}
    \label{eq:supp_projected_cost}
\end{equation}
With the unit line normal
\begin{equation}
    \mathbf{n}_i
    =
    \frac{1}{\|\mathbf{q}_i\|_2}
    \begin{bmatrix}
        -q_{iy}\\
        q_{ix}
    \end{bmatrix},
\end{equation}
the projector in Eq.~\eqref{eq:supp_projected_cost} satisfies
\begin{equation}
    \mathbf{I}_2
    -
    \frac{\mathbf{q}_i\mathbf{q}_i^{\top}}
         {\mathbf{q}_i^{\top}\mathbf{q}_i}
    =
    \mathbf{n}_i\mathbf{n}_i^{\top}.
\end{equation}
Therefore,
\begin{equation}
    \min_{\lambda_i\in\mathbb{R}}
    \|
        \mathbf{m}_i-\mathbf{c}-\lambda_i\mathbf{q}_i
    \|_2^2
    =
    \left[
        \mathbf{n}_i^{\top}(\mathbf{m}_i-\mathbf{c})
    \right]^2
    =
    e_i^2.
    \label{eq:supp_exact_metric}
\end{equation}
The projection above is unconstrained; the physical condition
$\lambda_i^{\ast}>0$ is imposed separately when validating robust hypotheses,
as described in the main paper.

Under a squared loss, the horizontal center can also be eliminated in closed
form. Define
\begin{equation}
    \mathbf{N}(\mathbf{R})
    =
    \begin{bmatrix}
        \mathbf{n}_1^{\top}\\
        \vdots\\
        \mathbf{n}_N^{\top}
    \end{bmatrix},
    \qquad
    \mathbf{b}(\mathbf{R})
    =
    \begin{bmatrix}
        \mathbf{n}_1^{\top}\mathbf{m}_1\\
        \vdots\\
        \mathbf{n}_N^{\top}\mathbf{m}_N
    \end{bmatrix}.
    \label{eq:supp_N_b}
\end{equation}
Then
\begin{equation}
    \sum_{i=1}^{N}e_i^2
    =
    \|\mathbf{b}-\mathbf{N}\mathbf{c}\|_2^2.
\end{equation}
If $\operatorname{rank}(\mathbf{N})=2$, its unique least-squares minimizer is
\begin{equation}
    \mathbf{c}^{\ast}(\mathbf{R})
    =
    (\mathbf{N}^{\top}\mathbf{N})^{-1}
    \mathbf{N}^{\top}\mathbf{b}.
    \label{eq:supp_center_closed_form}
\end{equation}
Thus, for the squared-loss incidence objective,
\begin{equation}
\begin{split}
    &\min_{\mathbf{R}\in\mathrm{SO}(3),\,\mathbf{c},\,\{\lambda_i\}}
    \sum_{i=1}^{N}
    \|
        \mathbf{m}_i-\mathbf{c}
        -\lambda_i\mathbf{q}_i(\mathbf{R})
    \|_2^2
    \\
    &\qquad =
    \min_{\mathbf{R}\in\mathrm{SO}(3)}
    \sum_{i=1}^{N}
    e_i^2
    \bigl(
        \mathbf{R},
        \mathbf{c}^{\ast}(\mathbf{R})
    \bigr).
    \label{eq:supp_profile_equivalence}
\end{split}
\end{equation}
This makes explicit that the latent scales and horizontal translation are
nuisance variables that can be profiled out, leaving rotation as the nonlinear
outer variable. The robust objective in the main paper uses the same analytic
elimination of the per-correspondence scales while profiling the
two-dimensional center under the chosen robust loss.

\section{Five-correspondence hypothesis size}
Before eliminating the latent scales, $N$ correspondences provide $2N$
scalar equations in $N+5$ unknowns: $N$ scales
$\{\lambda_i\}_{i=1}^{N}$, three rotation parameters, and the
two-dimensional horizontal center. The generic degree-of-freedom count
therefore gives the necessary condition
\begin{equation}
    2N \geq N+5,
\end{equation}
or equivalently $N\geq5$. After variable projection, the same count is
seen directly: each correspondence contributes one scalar incidence
constraint $e_i(\mathbf{R},\mathbf{c})=0$ on the five pose degrees of
freedom.

Accordingly, LDVP uses five correspondences to generate each pose
hypothesis. For a sampled set, rotation candidates are evaluated by the
structured search described in the main paper, while the horizontal center
and per-correspondence scales are eliminated analytically. Each resulting hypothesis is scored over all correspondences by MSAC.
Robust center profiling and continuous rotation refinement are then applied
only in the final refit on the selected consensus. The counting argument
motivates the five-correspondence hypothesis size; it does not guarantee a
unique or valid solution for every five-correspondence configuration.
\section{Planar Observability of the Free-Depth Incidence Geometry}
\label{supp:planar_observability}

This section derives Eq.~\eqref{eq:linearized_residual} and the planar
Jacobian dependency used in the main paper.

Let
\begin{equation}
    \mathbf{v}_i
    =
    \mathbf{R}^{\top}\mathbf{r}_i
    =
    \begin{bmatrix}
        q_{ix} & q_{iy} & v_{iz}
    \end{bmatrix}^{\top},
\end{equation}
and perturb the rotation locally as
\begin{equation}
    \mathbf{R}(\delta\boldsymbol{\omega})
    =
    \mathbf{R}
    \exp\!\left(-[\delta\boldsymbol{\omega}]_{\times}\right),
    \qquad
    \delta\boldsymbol{\omega}
    =
    \begin{bmatrix}
        \omega_x & \omega_y & \omega_z
    \end{bmatrix}^{\top}.
    \label{eq:supp_rotation_perturbation}
\end{equation}
To first order,
\begin{equation}
    \delta\mathbf{v}_i
    =
    \delta\boldsymbol{\omega}\times\mathbf{v}_i,
\end{equation}
so the horizontal components vary as
\begin{equation}
    \delta q_{ix}
    =
    \omega_yv_{iz}-\omega_zq_{iy},
    \qquad
    \delta q_{iy}
    =
    \omega_zq_{ix}-\omega_xv_{iz}.
    \label{eq:supp_delta_q}
\end{equation}

Recall the unnormalized incidence residual
\begin{equation}
    g_i
    =
    -q_{iy}(X_i-C_x)+q_{ix}(Y_i-C_y).
\end{equation}
At a noiseless solution, Eq.~\eqref{eq:ldvp_base} gives
\begin{equation}
    X_i-C_x=\lambda_iq_{ix},
    \qquad
    Y_i-C_y=\lambda_iq_{iy}.
    \label{eq:supp_horizontal_incidence}
\end{equation}
Differentiating $g_i$ and substituting
Eqs.~\eqref{eq:supp_delta_q}--\eqref{eq:supp_horizontal_incidence} gives
\begin{equation}
\begin{split}
    \delta g_i
    ={}&
    q_{iy}\delta C_x-q_{ix}\delta C_y
    \\
    &+
    h_i(q_{ix}\omega_x+q_{iy}\omega_y)
    \\
    &-
    \lambda_i(q_{ix}^2+q_{iy}^2)\omega_z,
\end{split}
    \label{eq:supp_delta_g}
\end{equation}
where
\begin{equation}
    h_i
    =
    Z_i-C_z
    =
    \lambda_i v_{iz}.
\end{equation}
After stacking all correspondences, let
$\mathbf{q}_x=[q_{1x},\ldots,q_{Nx}]^{\top}$,
$\mathbf{q}_y=[q_{1y},\ldots,q_{Ny}]^{\top}$, and
$\mathbf{h}=[h_1,\ldots,h_N]^{\top}$. The relevant Jacobian columns are
\begin{equation}
    \mathbf{J}_{C_x}=\mathbf{q}_y,
    \qquad
    \mathbf{J}_{C_y}=-\mathbf{q}_x,
\end{equation}
and
\begin{equation}
    \mathbf{J}_{\omega_x}
    =
    \mathbf{h}\odot\mathbf{q}_x,
    \qquad
    \mathbf{J}_{\omega_y}
    =
    \mathbf{h}\odot\mathbf{q}_y.
    \label{eq:supp_jacobian_columns}
\end{equation}

If all matched points lie on a common horizontal plane, then
$h_i=H$ for every $i$. Equation~\eqref{eq:supp_jacobian_columns} becomes
\begin{equation}
    \mathbf{J}_{\omega_x}
    =
    -H\mathbf{J}_{C_y},
    \qquad
    \mathbf{J}_{\omega_y}
    =
    H\mathbf{J}_{C_x}.
    \label{eq:supp_planar_dependency}
\end{equation}
Hence the two tilt perturbations are first-order coupled with horizontal
translation. After the horizontal center is profiled out, these directions do
not provide independent first-order rotational constraints.

For a non-planar scene, write $h_i=H+\Delta h_i$. Then
\begin{equation}
    \mathbf{J}_{\omega_x}
    =
    -H\mathbf{J}_{C_y}
    +
    \Delta\mathbf{h}\odot\mathbf{q}_x,
    \qquad
    \mathbf{J}_{\omega_y}
    =
    H\mathbf{J}_{C_x}
    +
    \Delta\mathbf{h}\odot\mathbf{q}_y,
    \label{eq:supp_relief_dependency}
\end{equation}
where
$\Delta\mathbf{h}=[\Delta h_1,\ldots,\Delta h_N]^{\top}$.
The additional terms cannot, in general, be represented by a single
horizontal translation, explaining why height variation restores independent
constraints to the free-depth formulation.

Finally, this local limitation is shared by OPE and AH. From
Eq.~\eqref{eq:supp_threeway_relation}, let
$s_i=\|\mathbf{q}_i\|_2>0$. At a noiseless solution, $e_i=0$, so
differentiating the scaled residual relation gives
\begin{equation}
    \delta r_i^{\mathrm{OP}}
    =
    s_i\,\delta e_i,
    \qquad
    \delta r_i^{\mathrm{AH}}
    =
    -s_i\,\delta e_i.
\end{equation}
Therefore, after stacking the correspondences,
\begin{equation}
    \mathbf{J}_{\mathrm{OP}}
    =
    \mathbf{D}_s\mathbf{J}_{\mathrm{LDVP}},
    \;
    \mathbf{J}_{\mathrm{AH}}
    =
    -\mathbf{D}_s\mathbf{J}_{\mathrm{LDVP}},
    \;
    \mathbf{D}_s
    =
    \operatorname{diag}(s_1,\ldots,s_N).
\end{equation}
Since $\mathbf{D}_s$ is nonsingular,
\begin{equation}
    \operatorname{null}(\mathbf{J}_{\mathrm{OP}})
    =
    \operatorname{null}(\mathbf{J}_{\mathrm{AH}})
    =
    \operatorname{null}(\mathbf{J}_{\mathrm{LDVP}}).
    \label{eq:supp_common_nullspace}
\end{equation}
The planar weakness is therefore a property of the common DSM-free
orthographic--perspective incidence geometry, rather than a numerical artifact
of the LDVP parameterization.

}

%% file: main.bib
@article{Sekii2013,
  title={6-DOF Camera Pose Estimation Using Reference Points on an Aerial Image without Altitude Information},
  author={Taiki Sekii and Tomokazu Sato and Hideyuki Kume and Naokazu Yokoya},
  journal={IPSJ Transactions on Computer Vision and Applications},
  volume={5},
  number={ },
  pages={134-142},
  year={2013},
  doi={10.2197/ipsjtcva.5.134}
}

@article{DA3,
  title={Depth anything 3: Recovering the visual space from any views},
  author={Lin, Haotong and Chen, Sili and Liew, Junhao and Chen, Donny Y and Li, Zhenyu and Shi, Guang and Feng, Jiashi and Kang, Bingyi},
  journal={arXiv preprint arXiv:2511.10647},
  year={2025}
}

@article{ye2026deco,
  title={DECO: Depth-Guided Co-Visibility Reasoning for Low-Altitude UAV Visual Localization},
  author={Yibin Ye and Xichao Teng and Shuo Chen and  Xiaokai Song, Dongdong Guan and Qifeng Yu and Zhang Li},
  journal={arXiv preprint arXiv:2608.22289},
  year={2026}
}

@article{depthpro,
  title={Depth pro: Sharp monocular metric depth in less than a second},
  author={Bochkovskii, Aleksei and Delaunoy, Ama{\~a}{\c{G}}l and Germain, Hugo and Santos, Marcel and Zhou, Yichao and Richter, Stephan R and Koltun, Vladlen},
  journal={arXiv preprint arXiv:2410.02073},
  year={2024}
}

@article{piccinelli2025unidepthv2,
  author={Piccinelli, Luigi and Sakaridis, Christos and Yang, Yung-Hsu and Segu, Mattia and Li, Siyuan and Abbeloos, Wim and Van Gool, Luc},
  journal={IEEE Transactions on Pattern Analysis and Machine Intelligence}, 
  title={UniDepthV2: Universal Monocular Metric Depth Estimation Made Simpler}, 
  year={2026},
  volume={48},
  number={3},
  pages={2354-2367}}

@article{zhao2023aliked,
  title   = {ALIKED: A Lighter Keypoint and Descriptor Extraction Network via Deformable Transformation},
  author  = {Zhao, Xiaoming and Wu, Xingming and Chen, Weihai and Chen, Peter C. Y. and Xu, Qingsong and Li, Zhengguo},
  journal = {IEEE Transactions on Instrumentation and Measurement},
  volume  = {72},
  pages   = {1--16},
  year    = {2023}
}

@article{collins2014infinitesimal,
  title={Infinitesimal Plane-Based Pose Estimation},
  author={Collins, Toby and Bartoli, Adrien},
  journal={International Journal of Computer Vision},
  volume={109},
  number={3},
  pages={252--286},
  year={2014},
  doi={10.1007/s11263-014-0725-5}
}

@article{torr2000mlesac,
  title={MLESAC: A New Robust Estimator with Application to Estimating Image Geometry},
  author={Torr, Philip H. S. and Zisserman, Andrew},
  journal={Computer Vision and Image Understanding},
  volume={78},
  number={1},
  pages={138--156},
  year={2000},
  doi={10.1006/cviu.1999.0832}
}

@article{DISK,
  title={DISK: Learning local features with policy gradient},
  author={Tyszkiewicz, Micha{\l} and Fua, Pascal and Trulls, Eduard},
  journal={Advances in neural information processing systems},
  volume={33},
  pages={14254--14265},
  year={2020}
}

@misc{deng2026aeromap3d,
      title={AeroMap3D: Anchoring Monocular UAV 6-DoF Localization to Visual-Geometric-Semantic Map Priors}, 
      author={Zhiyun Deng and Luis Sentis},
      year={2026},
      eprint={2607.14009},
      archivePrefix={arXiv},
      primaryClass={cs.RO},
      url={https://arxiv.org/abs/2607.14009}, 
}

@article{Lepetit2009EPnP,
  author  = {Lepetit, Vincent and Moreno-Noguer, Francesc and Fua, Pascal},
  title   = {{EPnP}: An Accurate O(n) Solution to the {PnP} Problem},
  journal = {International Journal of Computer Vision},
  volume  = {81},
  number  = {2},
  pages   = {155--166},
  year    = {2009},
  doi     = {10.1007/s11263-008-0152-6}
}

@inproceedings{Larsson_2021_ICCV,
  author    = {Larsson, Viktor and Pollefeys, Marc and Oskarsson, Magnus},
  title     = {Orthographic-Perspective Epipolar Geometry},
  booktitle = {Proceedings of the IEEE/CVF International Conference on Computer Vision (ICCV)},
  pages     = {5570--5578},
  year      = {2021}
}

@article{Li2026AH,
  author  = {Li, Yineng and Zeng, Qinghua and Jin, Ziqi and An, Yuqi and Li, Rongbing},
  title   = {Monocular Visual Localization Method Based on Antisymmetric-Homography Epipolar Constraint Theory},
  journal = {Measurement},
  volume  = {266},
  pages   = {120502},
  year    = {2026},
  doi     = {10.1016/j.measurement.2026.120502}
}

@article{Nikolakopoulos2020AW3D30,
  author  = {Nikolakopoulos, Konstantinos G.},
  title   = {Accuracy Assessment of {ALOS AW3D30 DSM} and Comparison to {ALOS PRISM DSM} Created with Classical Photogrammetric Techniques},
  journal = {European Journal of Remote Sensing},
  volume  = {53},
  pages   = {39--52},
  year    = {2020},
  doi     = {10.1080/22797254.2020.1774424}
}

@inproceedings{rublee2011orb,
  title={ORB: An efficient alternative to SIFT or SURF},
  author={Rublee, Ethan and Rabaud, Vincent and Konolige, Kurt and Bradski, Gary},
  booktitle={2011 International Conference on Computer Vision (ICCV)},
  pages={2564--2571},
  year={2011},
  organization={IEEE}
}

@ARTICLE{li2025imu,
  author={Li, Yongfei},
  journal={IEEE Robotics and Automation Letters}, 
  title={IMU-Aided Geographic Pose Estimation Method for UAVs Using Satellite Imageries Matching}, 
  year={2025},
  volume={10},
  number={3},
  pages={2902-2909}}

@article{zhang2026visual,
  title={Visual localization in 3D maps: comparing point cloud, mesh, and NeRF representations},
  author={Zhang, Lintong and Tao, Yifu and Lin, Jiarong and Zhang, Fu and Fallon, Maurice},
  journal={Autonomous Robots},
  volume={50},
  number={1},
  pages={14},
  year={2026},
  publisher={Springer}
}

@article{moge2aerial,
  title={AerialMetric: Benchmarking and Adapting UAV Monocular Metric Depth Estimation in the Real World},
  author={Song, Zhongqiang and Chen, Guanying and Zhang, Yuqi and Zou, Yin and Fu, Chuanyu and Yuan, Zhiyuan and Huang, Chuan and Cui, Shuguang and Cao, Xiaochun},
  journal={arXiv preprint arXiv:2606.29716},
  year={2026}
}

@article{moge2,
  title={Moge-2: Accurate monocular geometry with metric scale and sharp details},
  author={Wang, Ruicheng and Xu, Sicheng and Dong, Yue and Deng, Yu and Xiang, Jianfeng and Lv, Zelong and Sun, Guangzhong and Tong, Xin and Yang, Jiaolong},
  journal={Advances in Neural Information Processing Systems},
  volume={38},
  pages={35928--35959},
  year={2026}
}

@article{li2026robust,
  title={Robust 6-DoF Absolute Visual Localization for UAVs via Deep Homography Estimation},
  author={Li, Ban and Geng, Jianghui and Zhang, Hongping and Wang, Feng},
  journal={IEEE Transactions on Geoscience and Remote Sensing},
  year={2026},
  publisher={IEEE}
}

@article{wang2024accurate,
  title={Accurate vision-enabled UAV location using feature-enhanced transformer-driven image matching},
  author={Wang, Haoyang and Zhou, Fuhui and Wu, Qihui},
  journal={IEEE Transactions on Instrumentation and Measurement},
  volume={73},
  pages={1--11},
  year={2024},
  publisher={IEEE}
}

@inproceedings{Ye_2026_CVPR,
  author    = {Ye, Yibin and Teng, Xichao and Chen, Shuo and Liu, Leqi and Wang, Kun and Song, Xiaokai and Li, Zhang},
  title     = {Exploring the Best Way for UAV Visual Localization under Low-altitude Multi-view Observation Condition: A Benchmark},
  booktitle = {Proceedings of the IEEE/CVF Conference on Computer Vision and Pattern Recognition (CVPR) Findings},
  pages     = {1731--1741},
  year      = {2026}
}

@inproceedings{dhaouadi2025ortholoc,
  title     = {OrthoLoC: {UAV} 6-DoF Localization and Calibration Using Orthographic Geodata},
  author    = {Dhaouadi, Oussema and Marin, Riccardo and Meier, Johannes Michael and Kaiser, Jacques and Cremers, Daniel},
  booktitle = {Advances in Neural Information Processing Systems},
  year      = {2025}
}

@article{zhang2025GeoRVLF,
title={GeoRVLF: A Robust Drone-Satellite Visual Geo-Localization Framework for Small Unmanned Aerial Vehicle Platforms},
author={Zhongyuan Zhang and Jinkui Chu and Tao Song and Jikun Guo and Ran Zhang and Jinshan Li},
journal={IEEE Robotics and Automation Letters},
issue={7},
pages={7380-7387},
year={2025},
}

@misc{LoD-Loc,
      title={LoD-Loc: Aerial Visual Localization using LoD 3D Map with Neural Wireframe Alignment}, 
      author={Juelin Zhu and Shen Yan and Long Wang and Shengyue Zhang and Yu Liu and Maojun Zhang},
      year={2024},
      eprint={2410.12269},
      archivePrefix={arXiv},
      url={https://arxiv.org/abs/2410.12269}, 
}

@inproceedings{GPS_jamming,
author = {Xue, Nian and Niu, Liang and Hong, Xianbin and Li, Zhen and Hoffaeller, Larissa and P\"{o}pper, Christina},
title = {DeepSIM: GPS Spoofing Detection on UAVs using Satellite Imagery Matching},
year = {2020},
publisher = {Association for Computing Machinery},
address = {New York, NY, USA},
url = {https://doi.org/10.1145/3427228.3427254},
doi = {10.1145/3427228.3427254},
booktitle = {Proceedings of the 36th Annual Computer Security Applications Conference},
pages = {304–319},
numpages = {16},
location = {Austin, USA},
series = {ACSAC '20}
}

@inproceedings{detone18superpoint,
  author    = {Daniel DeTone and
               Tomasz Malisiewicz and
               Andrew Rabinovich},
  title     = {SuperPoint: Self-Supervised Interest Point Detection and Description},
  booktitle = {CVPRw},
  year      = {2018},
}

@inproceedings{lightglue,
  author    = {Philipp Lindenberger and
               Paul-Edouard Sarlin and
               Marc Pollefeys},
  title     = {{LightGlue: Local Feature Matching at Light Speed}},
  booktitle = {ICCV},
  year      = {2023}
}

@ARTICLE{SIFT,
  author={D. G. Lowe},
  journal={Int. J. Comput. Vis.}, 
  title={Distinctive image features from scale-invariant keypoints}, 
  year={2004},
  volume={60},
  number = {2},
  pages={91–110},
}

@inproceedings{toft2020single,
  title={Single-image depth prediction makes feature matching easier},
  author={Toft, Carl and Turmukhambetov, Daniyar and Sattler, Torsten and Kahl, Fredrik and Brostow, Gabriel J},
  booktitle={European conference on computer vision},
  pages={473--492},
  year={2020},
  organization={Springer}
}

@inproceedings{wang2023guiding,
  title={Guiding local feature matching with surface curvature},
  author={Wang, Shuzhe and Kannala, Juho and Pollefeys, Marc and Barath, Daniel},
  booktitle={Proceedings of the IEEE/CVF International Conference on Computer Vision},
  pages={17981--17991},
  year={2023}
}

@article{huang2023deep,
  title={Deep image registration with depth-aware homography estimation},
  author={Huang, Chenwei and Pan, Xiong and Cheng, Jingchun and Song, Jiajie},
  journal={IEEE Signal Processing Letters},
  volume={30},
  pages={6--10},
  year={2023},
  publisher={IEEE}
}

@inproceedings{karpur2024lfm,
  title={Lfm-3d: Learnable feature matching across wide baselines using 3d signals},
  author={Karpur, Arjun and Perrotta, Guilherme and Martin-Brualla, Ricardo and Zhou, Howard and Araujo, Andre},
  booktitle={2024 International Conference on 3D Vision (3DV)},
  pages={11--20},
  year={2024},
  organization={IEEE}
}

@INPROCEEDINGS{LiftFeat,
  author={Liu, Yepeng and Lai, Wenpeng and Zhao, Zhou and Xiong, Yuxuan and Zhu, Jinchi and Cheng, Jun and Xu, Yongchao},
  booktitle={2025 IEEE International Conference on Robotics and Automation (ICRA)}, 
  title={LiftFeat: 3D Geometry-Aware Local Feature Matching}, 
  year={2025},
  volume={},
  number={},
  pages={11714-11720},
  doi={10.1109/ICRA55743.2025.11127853}
}

@article{edstedt2024roma,
title={{RoMa: Robust Dense Feature Matching}},
author={Edstedt, Johan and Sun, Qiyu and Bökman, Georg and Wadenbäck, Mårten and Felsberg, Michael},
journal={IEEE Conference on Computer Vision and Pattern Recognition},
year={2024}
}

@inproceedings{surf,
  title={Surf: Speeded up robust features},
  author={Bay, Herbert and Tuytelaars, Tinne and Van Gool, Luc},
  booktitle={ECCV},
  pages={404--417},
  year={2006},
  organization={Springer}
}

@inproceedings{DKM,
title={{DKM}: Dense Kernelized Feature Matching for Geometry Estimation},
author={Edstedt, Johan and Athanasiadis, Ioannis and Wadenbäck, Mårten and Felsberg, Michael},
booktitle={CVPR},
year={2023}
}

@article{roma,
title={{RoMa: Robust Dense Feature Matching}},
author={Edstedt, Johan and Sun, Qiyu and Bökman, Georg and Wadenbäck, Mårten and Felsberg, Michael},
journal={CVPR},
year={2024}
}

@inproceedings{chen2021real,
  title={Real-time geo-localization using satellite imagery and topography for unmanned aerial vehicles},
  author={Chen, Shuxiao and Wu, Xiangyu and Mueller, Mark W and Sreenath, Koushil},
  booktitle={2021 IEEE/RSJ International Conference on Intelligent Robots and Systems (IROS)},
  pages={2275--2281},
  year={2021},
  organization={IEEE}
}

@inproceedings{sun2021loftr,
  title={LoFTR: Detector-Free Local Feature Matching with Transformers},
  author={Jiaming Sun and Zehong Shen and Yuang Wang and Hujun Bao and Xiaowei Zhou},
  booktitle = {CVPR},
  year={2021}
}

@article{chang2023review,
  title={A review of UAV autonomous navigation in GPS-denied environments},
  author={Chang, Yingxiu and Cheng, Yongqiang and Manzoor, Umar and Murray, John},
  journal={Robotics and Autonomous Systems},
  pages={104533},
  year={2023},
  publisher={Elsevier}
}

@inproceedings{goforth2019gps,
  title={GPS-denied UAV localization using pre-existing satellite imagery},
  author={Goforth, Hunter and Lucey, Simon},
  booktitle={2019 International conference on robotics and automation (ICRA)},
  pages={2974--2980},
  year={2019},
  organization={IEEE}
}

@article{he2024leveraging,
  title={Leveraging Map Retrieval and Alignment for Robust UAV Visual Geo-Localization},
  author={He, Mengfan and Liu, Jiacheng and Gu, Pengfei and Meng, Ziyang},
  journal={IEEE Transactions on Instrumentation and Measurement},
  year={2024},
  publisher={IEEE}
}

@article{ye2024coarse,
  title={A coarse-to-fine visual geo-localization method for GNSS-denied UAV with oblique-view imagery},
  author={Ye, Qin and Luo, Junqi and Lin, Yi},
  journal={ISPRS Journal of Photogrammetry and Remote Sensing},
  volume={212},
  pages={306--322},
  year={2024},
  publisher={Elsevier}
}

@inproceedings{Wu2023uav4l,
    title={UAVD4L: A Large-Scale Dataset for UAV 6-DoF Localization}, 
    author={Rouwan Wu and Xiaoya Cheng and Juelin Zhu and Xuxiang Liu and Maojun Zhang and Shen Yan},
    booktitle={International Conference on 3D Vision (3DV)},
    year={2024}
}
